\documentclass[sigconf]{acmart}

\AtBeginDocument{
  }  

\setcopyright{acmlicensed}
\copyrightyear{2026}
\acmYear{2026}
\acmDOI{XXXXXXX.XXXXXXX}

\acmConference[KDD '27]{The 33rd ACM SIGKDD Conference on Knowledge Discovery and Data Mining}{August 01--05, 2027}{San Jose, CA, USA}

\acmISBN{978-1-4503-XXXX-X/2018/06}

\acmSubmissionID{19}

\usepackage{booktabs}
\usepackage{multirow}
\usepackage{makecell}
\usepackage{graphicx}
\usepackage[table]{xcolor}
\usepackage{enumitem}
\usepackage{subcaption}
\usepackage{tcolorbox}
\usepackage[normalem]{ulem}
\usepackage{pifont}
\usepackage{etoolbox}

\definecolor{tabgroup}{RGB}{240,248,246}
\definecolor{tabhead}{RGB}{236,245,252}
\definecolor{projectpink}{HTML}{FF3B8D}

\newcommand{\cmark}{\ding{51}}
\newcommand{\xmark}{\ding{55}}
\newcommand{\yxnote}[1]{\textcolor{blue}{#1}}

\makeatletter
\patchcmd{\@mkbibcitation}
  {\ifx\@subtitle\@empty. \else: \@subtitle. \fi}
  {}
  {}
  {}
\makeatother

\begin{document}

\title{Towards Unified Multimodal Misinformation Detection in Social Media:  A Benchmark Dataset and Baseline}

\author{Haiyang Li}
\email{l1228hy@gmail.com}
\affiliation{%
  \institution{Hefei University of Technology}
  \city{Hefei}
  \country{China}
}

\author{Yaxiong Wang}
\email{wangyx15@stu.xjtu.edu.cn}
\affiliation{%
  \institution{Hefei University of Technology}
  \city{Hefei}
  \country{China}
}

\author{Shengeng Tang}
\email{tangsg@hfut.edu.cn}
\affiliation{%
  \institution{Hefei University of Technology}
  \city{Hefei}
  \country{China}
}

\author{Yuchen Zhang}
\email{yczhang@stu.edu.cn}
\affiliation{%
  \institution{Xi'an Jiaotong University}
  \city{Xi'an}
  \country{China}
}

\author{Lianwei Wu}
\email{wlw@nwpu.edu.cn}
\affiliation{%
  \institution{Northwestern Polytechnical University}
  \city{Xi'an}
  \country{China}
}

\author{Lechao Cheng}
\email{chenglc@hfut.edu.cn}
\affiliation{%
  \institution{Hefei University of Technology}
  \city{Hefei}
  \country{China}
}

\author{Liu Liu}
\email{liuliu@hfut.edu.cn}
\affiliation{%
  \institution{Hefei University of Technology}
  \city{Hefei}
  \country{China}
}

\author{Chaofeng Dong}
\email{nautydcf@126.com}
\affiliation{%
  \institution{BIREN Technology}
  \city{Shanghai}
  \country{China}
}

\author{Zhun Zhong}
\email{zhunzhong007@gmail.com}
\affiliation{%
  \institution{Hefei University of Technology}
  \city{Hefei}
  \country{China}
}


\renewcommand{\shortauthors}{Li et al.}

\begin{abstract}
Detecting deceptive multimodal content on social media has become an increasingly important problem. Two major types of deception dominate: human-crafted misinformation (e.g., rumors and misleading posts) and AI-generated content produced by image synthesis models or vision–language models (VLMs). However, these two types are usually addressed as separate tasks. Consequently, existing models are often specialized for only one type of fake content. In real deployments, however, the fake-content type of an incoming multimodal post is typically unknown, which limits the practicality of such specialized systems. To study this setting, we build OmniFake, a benchmark with 98K samples that combines human-curated misinformation from existing resources with newly created AI-generated examples. To address this new task, we propose Unified Multimodal Fake Content Detection (UMFDet), a framework designed to handle both types of deception. UMFDet builds on a VLM backbone augmented with a Category-aware Mixture-of-Experts (CMoE) adapter to capture category-specific cues.  We further introduce an Expert-wise Discriminative Regularization to enforce intra-expert compactness.  In addition, cross-modal consistency alignment is proposed to improve the perceptual capability of experts for handling different deception types. Experiments show that UMFDet consistently outperforms competitive specialized baselines across both deception types. Our dataset and code are publicly available at our
\href{https://0112hy.github.io/OmniFakeWeb/}
{\textcolor{projectpink}{\textbf{project page}}}.
\end{abstract}

\ccsdesc[500]{Information systems~Data mining}
\ccsdesc[500]{Computing methodologies~Machine learning}
\ccsdesc[300]{Computing methodologies~Computer vision}
\ccsdesc[300]{Computing methodologies~Natural language processing}

\received{20 February 2007}
\received[revised]{12 March 2009}
\received[accepted]{5 June 2009}

\keywords{
Multimodal DeepFake Detection, Fake News Detection, Manipulation Recognition
}

\subtitle{%
  \texorpdfstring{%
    \raisebox{-0.18\height}{%
      \includegraphics[height=1.1em]{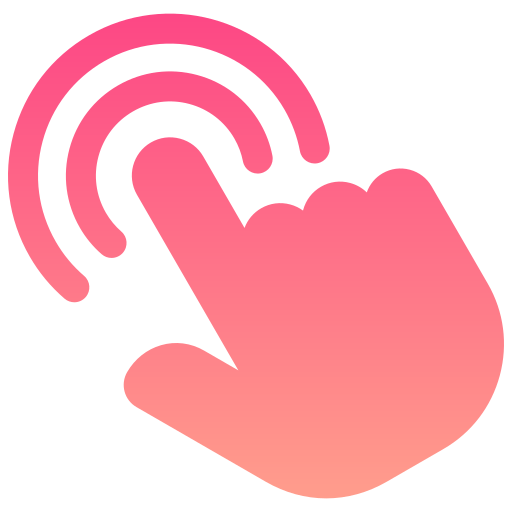}%
    }%
    \hspace{0.3em}%
    \href{https://0112hy.github.io/OmniFakeWeb/}{%
      \textcolor{projectpink}{%
        \bfseries Project Page: https://0112hy.github.io/OmniFakeWeb/%
      }%
    }%
  }{Project Page: https://0112hy.github.io/OmniFakeWeb/}%
}

\maketitle

\section{Introduction}


The proliferation of multimodal misinformation on social media platforms has emerged as a public security concern~\cite{liu2024mmfakebench, tahmasebi2024multimodal, guo2025each}. In response, detecting deceptive content such as rumors, misleading claims, and fabricated information has attracted considerable research interest in recent years~\cite{hu2024bad, zhang2024reinforced, wu2024fake,10.1145/3770854.3780211}. Mainstream misinformation can be broadly categorized into two types: human-crafted misinformation and AI-synthesized content. Human-crafted misinformation is intentionally created by individuals, often with motives such as gaining followers on social platforms. In contrast, AI-generated content is produced using generative AI tools, including image synthesis models, Large Language Models (LLMs), and Multimodal Large Language Models (MLLMs). The rise of AI-synthesized misinformation represents a growing challenge in the era of advanced deep learning systems~\cite{guo2024detective, chen2025study}.

Although both human-crafted misinformation and AI-generated content belong to misinformation, they are often studied in isolation. Researchers in natural language processing (NLP) have extensively studied human-crafted deceptive content, such as human-written rumors and misleading information, with a long-standing tradition of fake news detection in social media~\cite{10.1145/3637528.3672024, cui2024propagation, jiang2025epidemiology}. On the other hand, the computer vision community has focused primarily on detecting synthetic multimodal media generated by AI models, including deepfakes and other artificially produced visual content~\cite{yan2024df40,cheng2024can,zhang2025asap}. However, in real-world social platforms, both types of misinformation frequently coexist. A straightforward alternative is to maintain two separate detectors, each specialized for one source, and aggregate their predictions at inference time. Nevertheless, such a pipeline implicitly assumes source-specific dispatch and incurs additional computational overhead, making it less desirable for practical deployment. Moreover, existing approaches often struggle when facing fine-grained and diverse manipulation types from multiple sources, where the interaction between human-crafted misinformation and AI-generated content leads to significant performance degradation in real-world scenarios.
\begin{figure}[!t]
  \centering
  \includegraphics[width=\columnwidth]{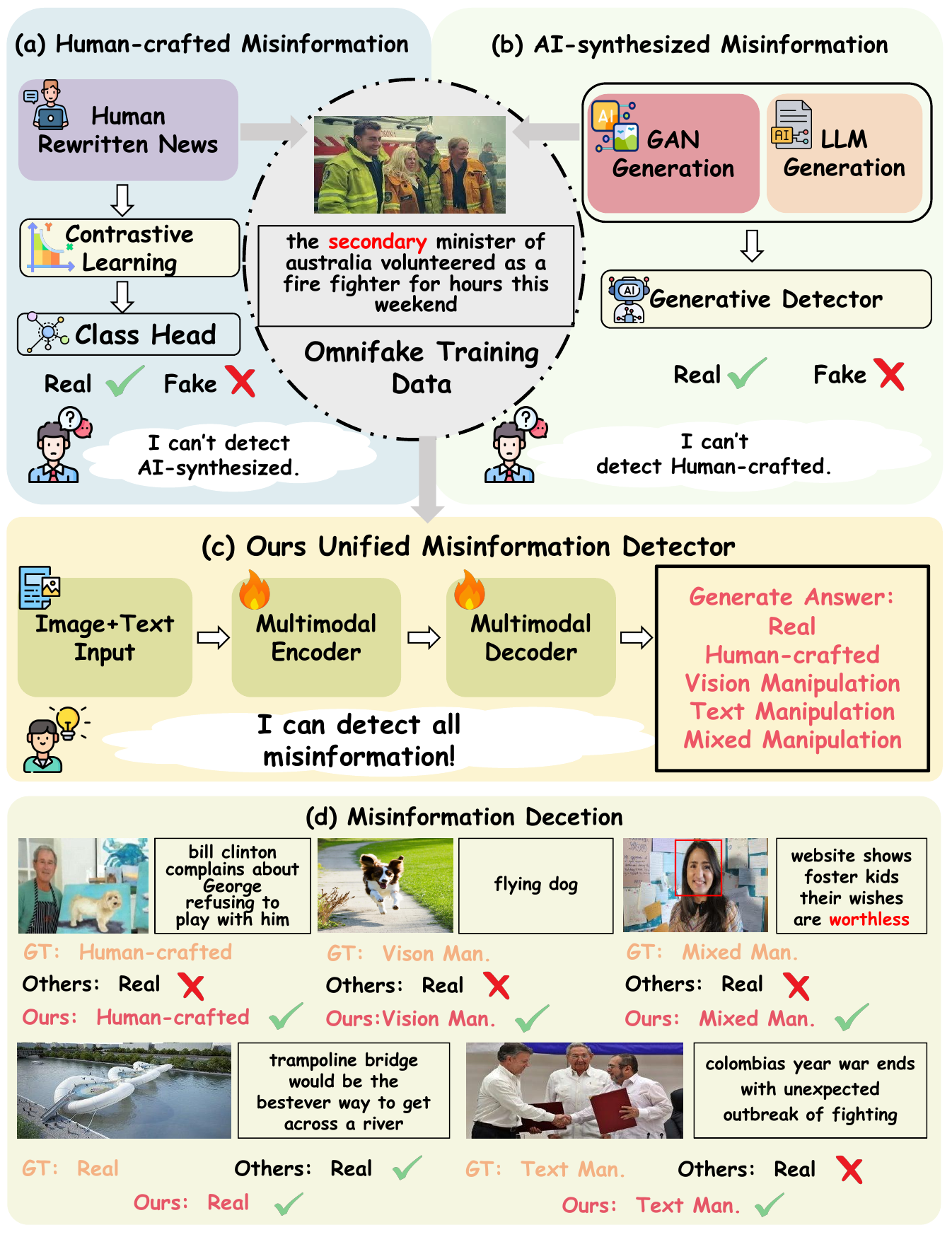} 
  \vspace{-0.5cm}
  \caption{Comparison between conventional detectors and our unified misinformation detector.}
  
  \label{fig:motivation}
  \vspace{-0.6cm}
\end{figure}

In response to this challenge, we aim to develop a unified framework capable of simultaneously detecting both human-crafted misinformation and AI-generated fake contents. To support this exploration, we construct a large-scale Omnibus Dataset for Multimodal News Deception (OmniFake). Specifically, we collect source samples from the r/Fakeddit~\cite{nakamura2019r} and VisualNews~\cite{nakamura2019r} dataset and select misleading examples to form the category of human-crafted misinformation. For the AI-generated category, we first choose a subset of real samples from r/Fakeddit and apply a diverse set of manipulation techniques including face swapping, background replacement, and textual rewriting \emph{etc.}, to simulate various forms of synthetic media. Through this process, We finally construct OmniFake, a dataset comprising 98,592 samples spanning five categories: \emph{Real}, \emph{Human-crafted}, \emph{Vision Manipulation}, \emph{Text Manipulation}, and \emph{Mixed Manipulation}.

\begin{figure*}[t]
  \centering

  \includegraphics[width=\textwidth]
  {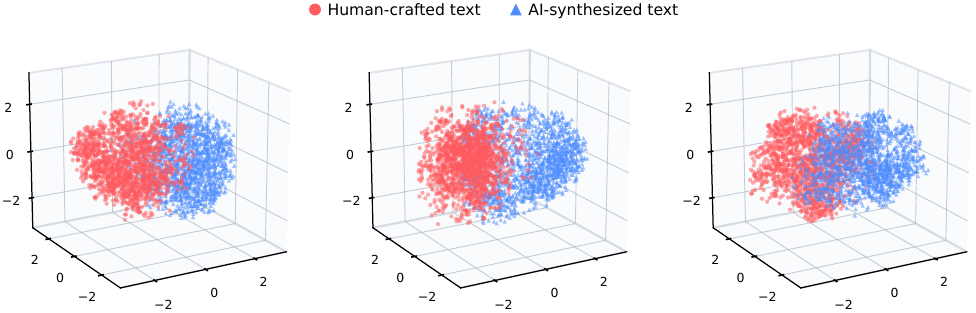}

  \begin{subfigure}[t]{0.32\textwidth}
    \centering
    \vspace{-0.3cm}
    \caption{BERT (Text-only)}
    \label{fig:vis_bert}
  \end{subfigure}
  \hfill
  \begin{subfigure}[t]{0.32\textwidth}
    \centering
    \vspace{-0.3cm}

    \caption{CLIP-ViT-B/32}
    \label{fig:vis_clip}
  \end{subfigure}
  \hfill
  \begin{subfigure}[t]{0.32\textwidth}
    \centering
    \vspace{-0.3cm}

    \caption{Qwen2.5-7B}
    \label{fig:vis_qwen}
  \end{subfigure}
\vspace{-0.4cm}
  \caption{Visualization of textual feature distributions across different
  backbones. We compare the representation spaces of
  \emph{Human-crafted text} and \emph{AI-synthesized text}.}
  \label{fig:text_3d_vis}
  \vspace{-0.2cm}
\end{figure*}

The OmniFake dataset presents two distinct yet deceptive forms of misinformation. Beyond the visual discrepancies inherent in human-crafted versus AI-generated news, the textual contents also exhibit considerable heterogeneity (see Fig.\ref{fig:text_3d_vis}). Motivated by these observations, we move beyond binary classification to treat these as separate categories for fine-grained recognition. Specifically, we propose the Unified Multimodal Fake Content Detection (UMFDet) framework. UMFDet leverages a Multimodal Large Language Model (MLLM) as its backbone and introduces a Category-aware Mixture-of-Experts (MoE) Adapter to tame the MLLM for misinformation detection. This adapter allocates specialized expert modules to identify specific types of deceptive content. To further enhance the model’s ability to accurately discern subtle cues of different types of misinformation, we enforce an Expert-wise Discriminative Regularization (EDR) computed on the routed sets: within each expert subspace, we estimate class prototypes and shrink intra-class variance while enlarging inter-class margins. Furthermore, a cross-modal consistency alignment learning is incorporated to bolster the experts' perceptual capability towards semantic correlations, effectively distinguishing between real news and human-crafted rumors through global consistency modeling.

In summary, we highlight our contributions as follows:
\begin{itemize}
\item \textbf{OmniFake dataset.}
We construct OmniFake, a large-scale benchmark that covers two prevalent
types of misinformation and enables unified evaluation across them.

\item \textbf{UMFDet framework.}
We propose UMFDet, a unified detection framework that combines a
category-aware MoE adapter, expert-wise discriminative regularization,
and cross-modal consistency alignment for this setting. Extensive
experiments demonstrate its effectiveness.

\item \textbf{Extensive experimental study on effectiveness and generalization.}
We conduct comprehensive evaluations on five datasets.
Our model outperforms competing methods in both effectiveness and generalization.

\end{itemize}







\begin{table*}[t]
\centering
\caption{Comparison with existing multimodal fake news datasets. Modalities: Text (Txt.) and Image (Img.). Category: Real (\textbf{R}), Human-crafted (\textbf{H}), AI-synthesized (\textbf{A}), and Mixed Manipulation (\textbf{M}). Mani.: Manipulated. PIG/E: Part image generation/edit; PTG/E: Part text generation/edit; FTG: Full text generation; FIG: Full image generation. LLM Threat: suitable for evaluating LLM risks. Img./Txt. BBox: availability of image and text bounding-box annotations. Tr.: available training set.}
\label{tab:datasets_comparison}
\vspace{-0.3cm}
\small
\setlength{\tabcolsep}{2.2pt}
\renewcommand{\arraystretch}{1.08}

\resizebox{\textwidth}{!}{%
\begin{tabular}{l cc c c c cc cc c c c c}
\toprule
\multirow{2}{*}{\textbf{Datasets}}
& \multicolumn{2}{c}{\textbf{Mod.}}
& \multirow{2}{*}{\textbf{Scale}}
& \multirow{2}{*}{\textbf{Category}}
& \multirow{2}{*}{\textbf{Lang.}}
& \multicolumn{2}{c}{\textbf{Mani.}}
& \multicolumn{2}{c}{\textbf{Img Type}}
& \multirow{2}{*}{\textbf{LLM}}
& \multirow{2}{*}{\shortstack[c]{\textbf{Img./Txt.}\\\textbf{BBox}}}
& \multirow{2}{*}{\textbf{Data Collection}}
& \multirow{2}{*}{\textbf{Tr.}} \\
\cmidrule(lr){2-3}
\cmidrule(lr){7-8}
\cmidrule(lr){9-10}

& \textbf{Txt.} & \textbf{Img.}
& & &
& \textbf{Txt.} & \textbf{Img.}
& \textbf{Face} & \textbf{Oth.}
& & & & \\
\midrule

Twitter~\cite{boididou2015verifying}
& \cmark & \cmark
& 18K & R, H & EN
& \xmark & \xmark
& \xmark & \xmark
& \xmark
& \xmark
& Social platform
& \cmark \\

Pheme~\cite{zubiaga2016learning}
& \cmark & \xmark
& 29K & R, H & EN
& \xmark & \xmark
& \xmark & \xmark
& \xmark
& \xmark
& Social platform
& \cmark \\

Weibo~\cite{nan2021mdfend}
& \cmark & \cmark
& 11K & R, H & CH
& \xmark & \xmark
& \xmark & \xmark
& \xmark
& \xmark
& Social platform
& \cmark \\

NewsCLIPpings~\cite{luo2021newsclippings}
& \cmark & \cmark
& 85K & R, A & EN
& \xmark & \xmark
& \xmark & \xmark
& \xmark
& \xmark
& Social platform
& \cmark \\

DGM4~\cite{shao2024detecting}
& \cmark & \cmark
& 50K & R, A & EN
& \cmark & \cmark
& \cmark & \xmark
& \xmark
& \cmark
& Soc.; PIE; PTE
& \cmark \\

MDSM~\cite{zhang2025coherence}
& \cmark & \cmark
& 441K & R, A & EN
& \cmark & \cmark
& \cmark & \xmark
& \xmark
& \xmark
& Soc.; PIE; PTE
& \cmark \\

MMFakeBench~\cite{liu2024mmfakebench}
& \cmark & \cmark
& 11K & \shortstack[c]{R, H, A} & EN
& \cmark & \cmark
& \xmark & \cmark
& \cmark
& \xmark
& \shortstack[c]{Soc.; PIG; PIE; FIG;\\PTE; PTG; FTG}
& \xmark \\

\midrule

\rowcolor{tabgroup}
\textbf{OmniFake (Ours)}
& \cmark & \cmark
& \textbf{98K}
& \textbf{\shortstack[c]{R, H, A, M}}
& EN
& \cmark & \cmark
& \cmark & \cmark
& \cmark
& \cmark
& \textbf{\shortstack[c]{Soc.; PIG; PIE; FIG;\\PTE; PTG; FTG}}
& \cmark \\

\bottomrule
\end{tabular}%
}
\vspace{-0.3cm}
\end{table*}

\section{Omnibus Dataset for Multimodal News Deception Detection}

We build OmniFake upon r/Fakeddit~\cite{nakamura2019r} and VisualNews~\cite{liu2020visualnews}. r/Fakeddit contains large-scale Reddit image--text pairs spanning 22 subreddits, while VisualNews provides professionally curated pairs from diverse news media and domains. Together, they offer over one million image--text pairs covering topics from politics to everyday events.

OmniFake contains five categories: \emph{Real}, \emph{Human-crafted}, \emph{Vision Manipulation}, \emph{Text Manipulation}, and \emph{Mixed Manipulation}. We select 49,034 authentic posts from r/Fakeddit and VisualNews for the \textbf{Real} category, and 24,726 misleading posts from r/Fakeddit for the \textbf{Human-crafted} category. The remaining samples are generated with modern image and language models to construct the \textbf{Vision Manipulation}, \textbf{Text Manipulation}, and \textbf{Mixed Manipulation} categories, as detailed below.

\subsection{Synthetic Multimodal News}
\begin{figure*}[t]
  \centering
  \includegraphics[width=\textwidth]{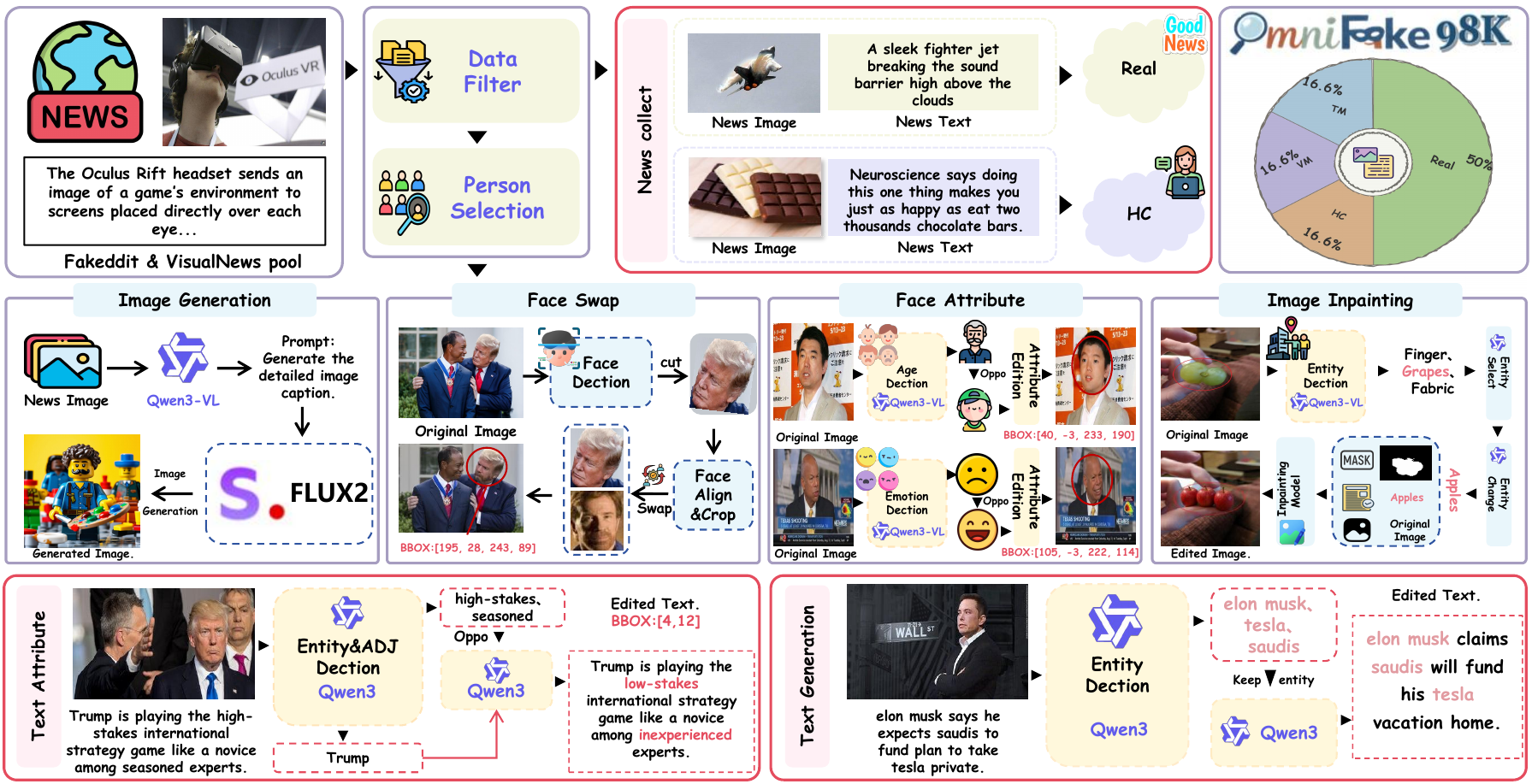}
  \vspace{-0.4cm}
  \caption{\textbf{Dataset construction pipeline.} We collect authentic and human-crafted posts from the source corpora to construct the \emph{Real} and \emph{Human-crafted} categories. For AI-synthesized misinformation, we generate diverse image-side manipulations, including full-image generation, local inpainting, face swapping, and facial attribute editing, as well as LLM-driven text generation and rewriting with controlled cross-modal semantics.}
  \label{fig:dataset_pipeline}
  \vspace{-0.3cm}
\end{figure*}

\textbf{Image Synthesis.}
For person-containing images (PCI), we apply face swapping and facial attribute editing; for non-person images (NPI), we use full-image generation and inpainting to produce diverse, realistic forgeries.

$\vartriangleright$ \emph{Face Swap Manipulation (PCI).} Face swapping poses risks on social media by undermining the credibility of manipulated images. We adopt SimSwap~\cite{chen2020simswap}, E4S~\cite{li2023e4s} and hyperswap~\cite{li2023e4s} for face swap. Given a source--target pair $\{I_s,I_t\}$, $I_s$ is sampled from source samples, while $I_t$ is randomly selected from CelebA\textendash HQ~\cite{zhu2022celebv}. We detect faces using DSFD~\cite{li2019dsfd}, select the highest-confidence pair, and perform landmark-based alignment and cropping to constrain the swap region. Conditioned on the identity embedding of $I_s$, the model transfers the source identity to the target face while preserving pose, expression, illumination, and context, then blends the generated face into $I_t$.

$\vartriangleright$ \emph{Face Attribute Manipulation (PCI).} Face attribute manipulation poses risks on social media by altering personal appearances and reducing image authenticity. We employ HFGI~\cite{wang2022high} to edit facial expressions, age, lips, and eyes. Given an image $I$, DSFD~\cite{li2019dsfd} detects the most prominent face, which is aligned and cropped into a face patch $F_c$ with landmarks $B$. HFGI then edits $F_c$ along the selected latent attribute direction to obtain $F_e$. For expression and age manipulation, Qwen3-VL~\cite{yang2025qwen3} estimates the original emotion or age to guide the corresponding reverse or targeted edit. The editing strength is controlled by $\alpha$, and the resulting face is seamlessly blended back into the original image using landmarks $B$.

$\vartriangleright$\emph{ Full Image Generation (NPI).} Image generation poses risks on social media by creating highly realistic synthetic images that may mislead viewers. To construct photorealistic AI-generated news images, we first use Qwen-3-VL~\cite{bai2025qwen2} to generate a detailed description $T$ of the original image. We then feed $T$ into SD-XL~\cite{podell2023sdxl} or FLUX~\cite{esser2024scalingrectifiedflowtransformers} to synthesize the manipulated image $\hat I_n$.

$\vartriangleright$\emph{Object/Background Replacement (NPI).} 
Local inpainting replacement is pervasive on social media, where forged news images alter only a small region to mislead viewers. Given an input image $I$, we use Qwen-3-VL~\cite{bai2025qwen2} to produce prompt phrases for the $r$ most salient objects, forming a set $\{n_j\}_{j=1}^{r}$ (with $r=3$ by default). We then randomly select one description $n_x$ as the target and feed $(n_x,I)$ into SAM2~\cite{ravi2024sam} to obtain the object mask $M$. For local tampering, we adopt an inpainting pipeline based on SDXL Inpainting~\cite{podell2023sdxl}: a language model converts the source object phrase $p_{\text{src}}=n_x$ into a modified phrase $p_{\text{mod}}$ by altering its type,color, or attribute while keeping unrelated context unchanged; the inpainting model takes $(I,M,p_{\text{mod}})$ and synthesizes the edited image.

\textbf{Text Fabrication.}
Text fabrication is challenging due to its easy creation and misleading potential. We construct two types of text manipulations while preserving named entities.

$\vartriangleright$ \emph{Pure Fake Text Generation.} Given a headline $T$, Qwen3~\cite{yang2025qwen3} extracts named entities $\mathcal{E}$ and rewrites all non-entity content into misleading text. If no entities are detected, the entire headline is rewritten, producing semantic-level misinformation.


$\vartriangleright$ \emph{Keyword/Phrase Distortion.} Qwen3~\cite{yang2025qwen3} extracts named entities $\mathcal{E}$ and candidate keywords $\mathcal{K}=\{k_j\}_{j=1}^{m}$ from $T$. It then replaces 2--3 keywords or short phrases, such as substituting adjectives with antonyms, while keeping entities unchanged. Each modified token or phrase and its position $P$ are recorded for localization.

\begin{figure*}[t]
\centering

\begin{subfigure}[t]{0.32\textwidth}
    \centering
    \includegraphics[width=\linewidth]{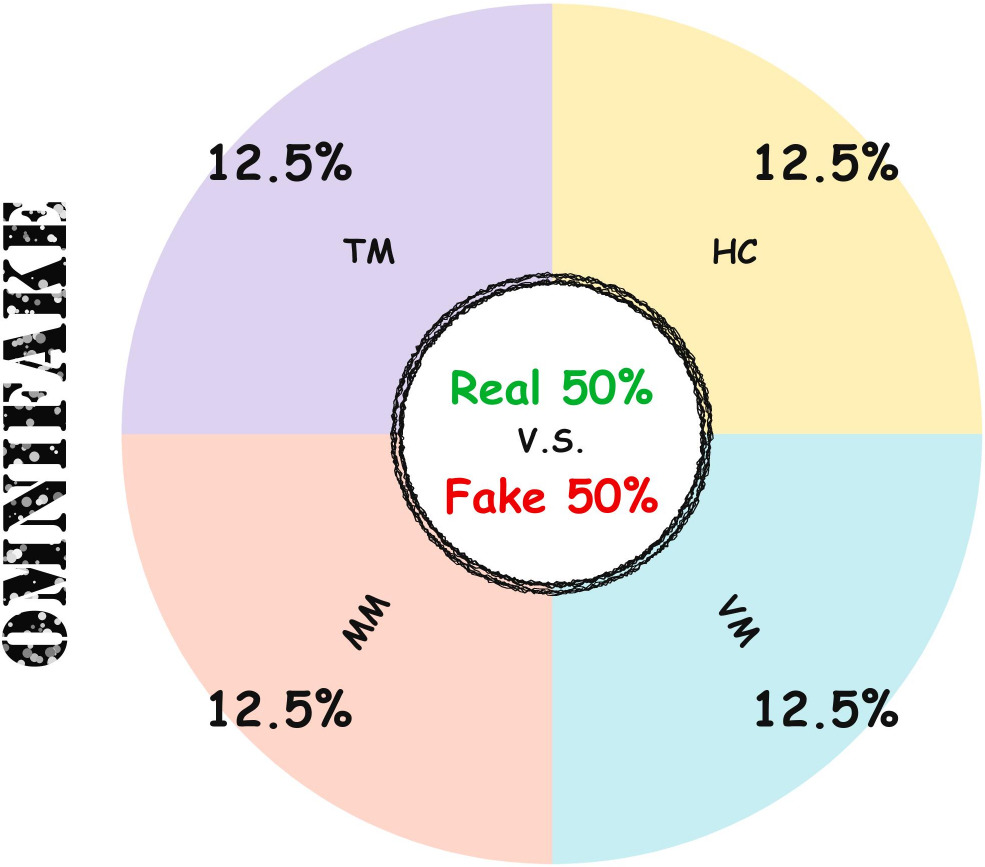}
    \caption{Category distribution.}
    \label{fig:omnifake_categories}
\end{subfigure}
\hfill
\begin{subfigure}[t]{0.28\textwidth}
    \centering
    \includegraphics[width=\linewidth]{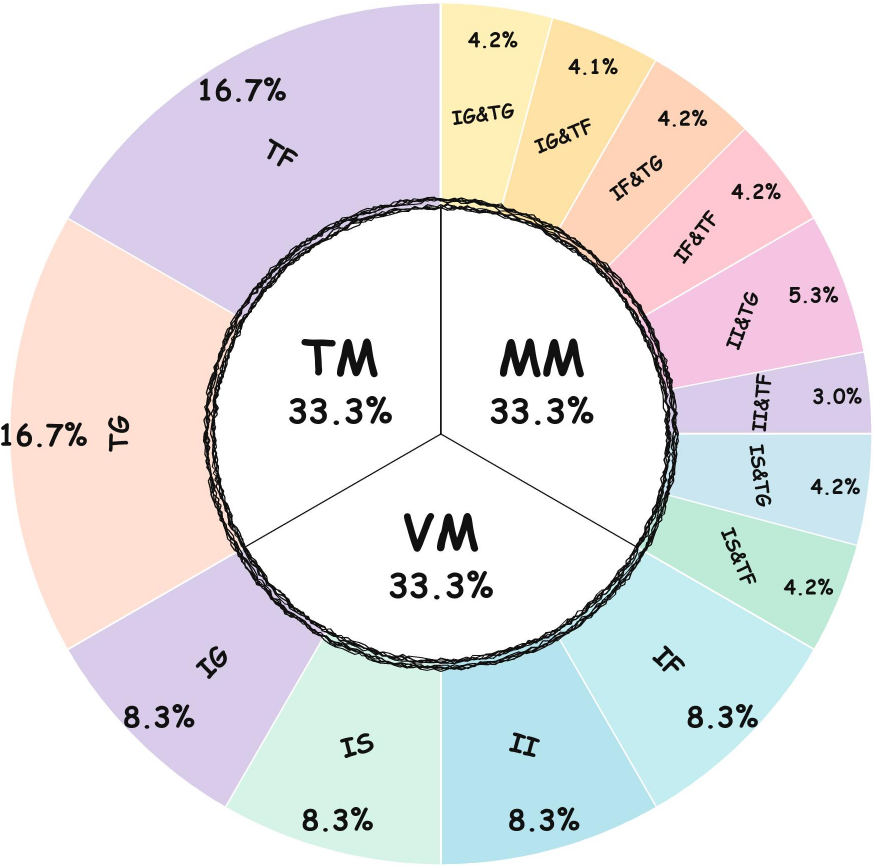}
    \caption{Fake subcategory distribution.}
    \label{fig:omnifake_fake_subcategories}
\end{subfigure}
\hfill
\begin{subfigure}[t]{0.36\textwidth}
    \centering
    \includegraphics[width=\linewidth]{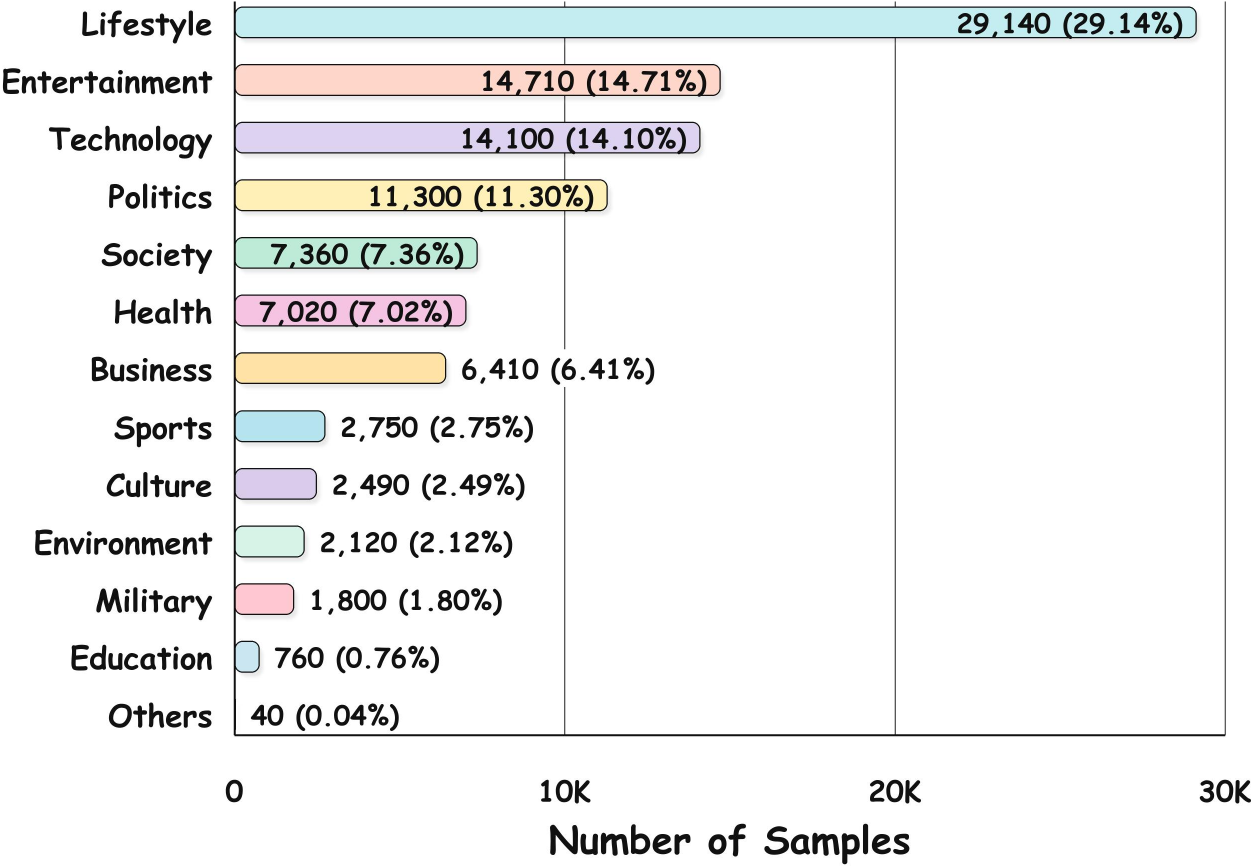}
    \caption{Domain distribution.}
    \label{fig:omnifake_domain_distribution}
\end{subfigure}
\vspace{-0.3cm}
\caption{Distributions of the five categories, manipulation types, and news domains, showing balanced classes, diverse manipulations, and broad topic coverage.}
\label{fig:omnifake_statistics}
\vspace{-0.3cm}
\end{figure*}

\subsection{Dataset Statistics}

With the above pipeline, we curate OmniFake, a multimodal fake-news benchmark for unified manipulation detection. As shown in Figure~\ref{fig:omnifake_statistics}, it has four key characteristics:





\begin{itemize}[leftmargin=1.2em,topsep=2pt,itemsep=1pt,parsep=0pt]
  \item \textbf{Comprehensive Coverage.} OmniFake unifies five categories in one unified benchmark: Real, Human-crafted, Vision Manipulation, Text Manipulation, and Mixed Manipulation.

  \item \textbf{Large Scale.} It contains 98,592 image--text pairs for robust evaluation of unified multimodal misinformation detection.

\item \textbf{Balanced Design.} Real and Fake samples are evenly distributed across sources and news domains, with carefully organized fine-grained fake categories for reliable evaluation.

  \item \textbf{Diverse Manipulations and Domains.} OmniFake covers full-image generation, local inpainting, face swapping, attribute editing, text rewriting, and eight mixed manipulation patterns across diverse real-world domains.
\end{itemize}

\subsection{Quality Evaluation}

\begin{figure*}[t]
  \centering
  \includegraphics[width=\textwidth]{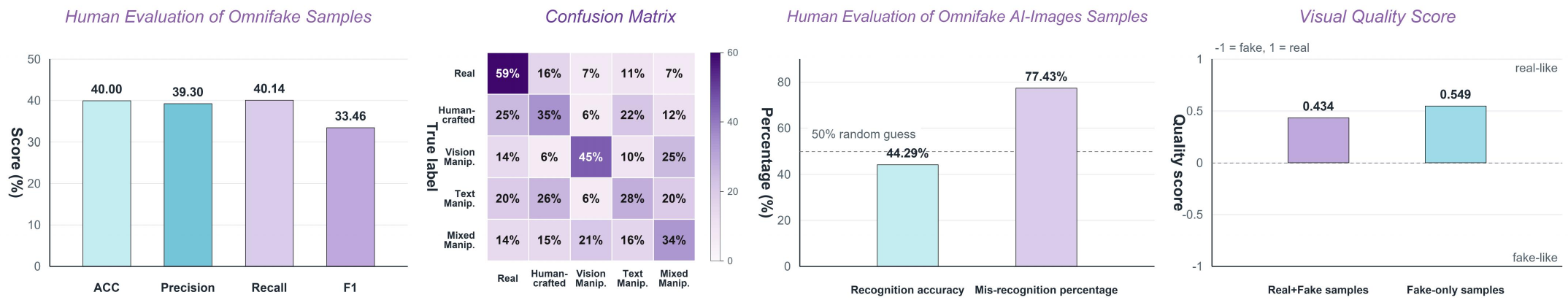}
  \vspace{-0.4cm}
  \caption{\textbf{Human evaluation on OmniFake.}
  The first two panels report human performance on OmniFake samples, including overall classification metrics and a row-normalized confusion matrix over five categories.
  The last two panels evaluate the visual quality of AI-generated images, showing recognition accuracy, mis-recognition percentage, and perceived quality scores.}
  \label{fig:human_eval}
  \vspace{-0.3cm}
\end{figure*}

We conduct a human audit on a randomly sampled subset of 600 OmniFake samples, evaluated by 8 master's students. As shown in Figure~\ref{fig:human_eval}, they achieve only 40.00\% accuracy and 33.46\% F1-score on the five-class task. The confusion matrix shows that human-crafted, text-manipulated, and mixed-manipulated samples are frequently confused, indicating the challenging and deceptive nature of OmniFake. For AI-image realism evaluation, we further assess 300 generated images, of which 77.43\% are mis-recognized as real, while fake-only images obtain a visual quality score of 0.549. These results validate the realism and deceptive quality of the AI-generated images in OmniFake, rather than merely reflecting generator artifacts.


\section{UMFDet Model}

\subsection{Architecture}

\begin{figure*}[t]
  \centering
  \includegraphics[width=\textwidth]{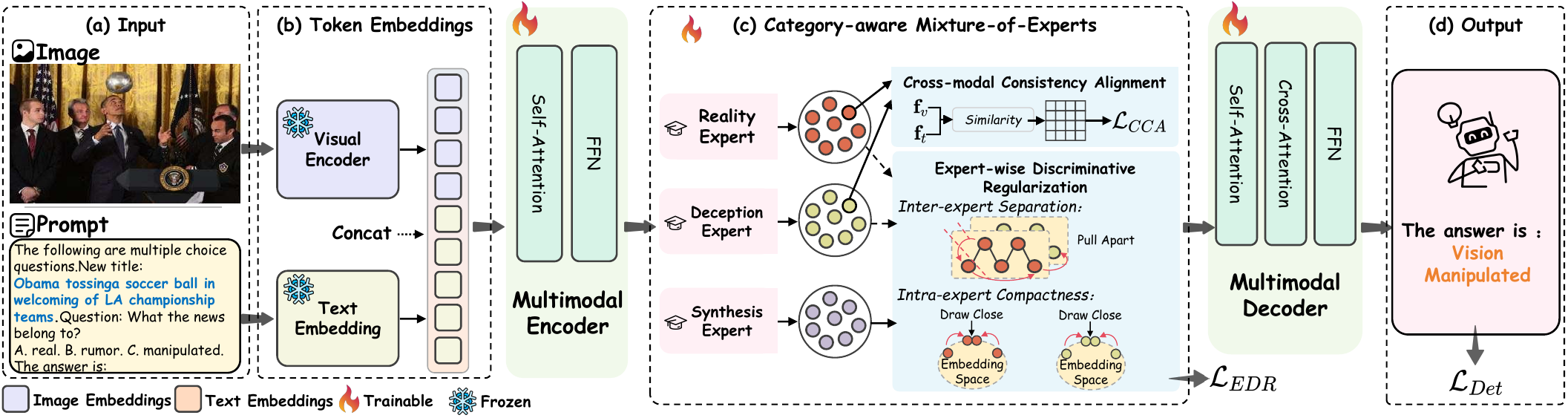}
  \caption{We propose UMFDet, a multimodal fake news detection framework
  based on CMoE, where EDR enhances expert discrimination and CCA improves
  cross-modal semantic alignment.}
  \label{fig:framework}
  \vspace{-0.3cm}
\end{figure*}

As shown in Figure~\ref{fig:framework}, we adopt Florence-2-0.7B~\cite{xiao2024florence} as the backbone for efficiency and effectiveness, and adapt it to multimodal misinformation detection with a Category-aware Mixture of Experts (CMoE). Following the MLLM paradigm, each image--title pair is formulated as a question-answering prompt. Visual and textual embeddings are fused by the multimodal encoder, processed by CMoE, and decoded to generate predictions. We further introduce Expert-wise Discriminative Regularization (EDR) to promote expert specialization and cross-modal consistency alignment (CCA) to capture image--text semantic correlations. The model is jointly optimized with the language modeling, EDR, and CCA loss.
\noindent\textbf{Image Embeddings.} Given input image $I$, we use DaViT~\cite{ding2022davit}as our visual encoder, generating flattened visual token embeddings $E_v\in \mathbb{R}^{N_v \times H}$, where $N_v$ represents the number of tokens and $H$ is the dimension of each token.
\vspace{-0.1cm}
\subsection{Instruction Design and Embedding} 

We build the instruction template as following:
\begin{equation}
T = [P_{\text{task}};\, P_{\text{opt}}(C);\, P_{\text{que}};\, P_{\text{resp}}], 
\end{equation}
where $P_{\text{task}}$ specifies the goal, $P_{\text{opt}}(C)$ lists the category set and brief definitions, $P_{\text{que}}$ define the question, and $P_{\text{resp}}$ declares the expected answer format. 

\noindent After tokenization and embedding, we obtain $E_t \in \mathbb{R}^{N_t \times H}$ for prompt representation. 
\vspace{-0.2cm}
\subsection{Category-aware Mixture-of-Expert}

Our Category-aware MoE (CMoE) uses three experts, each aligns with one primary category (real, human-crafted, AI-Tool synthesized). Given textual embeddings $E_t \in \mathbb{R}^{N_t \times H}$ and visual embeddings
$E_v \in \mathbb{R}^{N_v \times H}$, We fuse the concatenated multimodal inputs using a language encoder:
\begin{equation}
 X = L_\text{enc}([E_v; E_t]) \in \mathbb{R}^{(N_v+N_t)\times H},
\end{equation}

Subsequently, to capture category-specific features, we introduce a Category-aware Mixture-of-Experts (MoE) module composed of a router and 3 \yxnote{!} experts. In particular, we define three experts as: Reality Expert $\mathcal{E}_R$, Deception Expert $\mathcal{E}_D$, and Synthesis Expert $\mathcal{E}_S$, each specialized for a distinct category. Given a multimodal news representation $X$, the experts yield encodings $[\mathcal{E}_R(X), \mathcal{E}_D(X), \mathcal{E}_S(X)]$. Formally, each expert is a feed-forward network.
\begin{equation}
\begin{cases}
\mathcal{E}_e(X) = \mathrm{Dropout}(y)\, W_{\text{out}} + b_{\text{out}}, \\
y = u \odot v, \\
u = \mathrm{SiLU}(X W_{a} + b_{a}), \\
v= \operatorname{sigmoid}(X W_{b} + b_{b}), 
\end{cases}
\end{equation}
where all $W$ and $b$ are learnable parameters. The router network, implemented as a linear layer with softmax activation, then selects the optimal expert index $e$ as follows:
\begin{equation}
\begin{cases}
e = \operatorname*{arg\, max}_{e\in \{R, D, S\}} \{\mathcal{W}_R, \mathcal{W}_D, \mathcal{W}_S\}, \\
\left[\mathcal{W}_R, \mathcal{W}_D, \mathcal{W}_S\right] = \operatorname{Softmax}(W X + b),
\end{cases}
\end{equation}

Finally, we select $\mathcal{E}_e(X)$ and feed it to the language decoder for answer generation. Language modeling loss is used for training:
\begin{equation}
\mathcal{L}_{\text{Det}}
= - \sum_{t=1}^{|Y|}
\log p_{\theta}\!\big(y_t \, \big|\, y_{<t}, \, \mathcal{E}_e(X)\big),
\label{eq:cot_ce_final}
\end{equation}

\subsection{Expert-wise Discriminative Regularization}

For each expert, we expect it to specialize in a particular type of manipulation. To this end, we add a simple batch-wise regularizer on the CMoE outputs.

Let $\mathcal{S}_r = \{X_i\}_{i=1}^{|\mathcal{S}_r|}$ denote the set of samples routed to the {real expert} in the current mini-batch. Similarly, the sample sets corresponding to the {deception expert} and the {synthesis expert} are denoted as $\mathcal{S}_d$ and $\mathcal{S}_s$, respectively. To ease the constraint and stabilize the optimization, we dynamically select the two experts that receive the largest number of samples within each mini-batch. 
Formally, the selected sets are defined as:
\begin{equation}
  \{S_1, S_2\} = \operatorname{Top-2}(\{S_r, S_d, S_s\}, \text{key}=|S|), 
\end{equation}
Each expert subspace has a prototype and variance defined as:
\begin{align}
&\mu_k = \frac{1}{|\mathcal{S}_k|}\sum_{i=0}^{|\mathcal{S}_k|} X_i, \quad
\sigma_k^2 = \frac{1}{|\mathcal{S}_k|}\sum_{i=0}^{|\mathcal{S}_k|} \|f_i-\mu_k\|_2^2, \\
&d_{1, 2} = \|\mu_{1}-\mu_{2}\|_2^2, 
\end{align}
where $k\in \{1, 2\}$. Then, a clustering loss is defined to minimize within-expert variance and maximize inter-expert distance:
\begin{equation}
\mathcal{L}_{\text{EDR}} =
\frac{\sigma_{1}^{2}+\sigma_{2}^{2}}{d_{1, 2}+\varepsilon}, 
\end{equation}
where $\varepsilon>0$ is a small number for computation stability. EDR aims to encourage intra-category compactness within each expert subspace while enlarging the distance between prototypes of different expert subspaces, allowing each expert to focus on its category.

\subsection{Cross-modal Consistency Alignment}

In our practice, we found that the trained models still struggle to distinguish between authentic news and sophisticated human-crafted misinformation. 
Notably, the multimodal misinformation crafted by humans often repurposes images from unrelated sources, pairing them with misleading or out-of-context text. As a result, the semantic alignment between image and text in such deceptive content is typically weaker than that in authentic posts. Leveraging this prior, we introduce a Cross-modal Consistency Alignment (CCA) mechanism within the CMoE framework, which ensures that samples routed to different experts adhere to the underlying multimodal correlation prior.

Mathematically, let $X = [X_v; X_t]$ denote a sample from real or
deception expert set, where $X_v$ and $X_t$ are the visual and textual splits of the multimodal representation sequence $X$.
Subsequently, we define the target cross-modal alignment values as $y_{sim} = 1$ for real expert set and $y_{sim} = -1$ for the deception expert set ($\mathcal{C}_{hc}$). The loss is calculated using MSE over the respective subsets:
\begin{equation}
    \mathcal{L}_{CCA} = \sum_{X \in \mathcal{S}^{r}} \| \text{sim}(X_v, X_t) - 1 \|^2 + \sum_{X \in \mathcal{S}{^d}} \| \text{sim}(X_v, X_t) - (-1) \|^2,
\end{equation}
where $\text{sim}(\cdot,\cdot)$ is the cosine similarity function.

\begin{table*}[t]
\centering
\caption{Comparison of multimodal learning methods on OmniFake. AVG refers to the average performance across 5 classes.}
\label{tab:omnifake_main}
\vspace{-0.3cm}
\scriptsize
\setlength{\tabcolsep}{2.0pt}
\renewcommand{\arraystretch}{1.08}

\resizebox{\textwidth}{!}{%
\begin{tabular}{c l l ccc ccc ccc ccc ccc ccc}
\toprule
\multirow{2}{*}{\rotatebox[origin=c]{90}{\textbf{Setting}}} &
\multirow{2}{*}{\textbf{Method}} &
\multirow{2}{*}{\textbf{Venue}} &
\multicolumn{3}{c}{\textbf{Real}} &
\multicolumn{3}{c}{\textbf{Human-crafted}} &
\multicolumn{3}{c}{\textbf{Vision Manipulation}} &
\multicolumn{3}{c}{\textbf{Text Manipulation}} &
\multicolumn{3}{c}{\textbf{Mixed Manipulation}} &
\multicolumn{3}{c}{\textbf{AVG 5-class}} \\
\cmidrule(lr){4-6}
\cmidrule(lr){7-9}
\cmidrule(lr){10-12}
\cmidrule(lr){13-15}
\cmidrule(lr){16-18}
\cmidrule(lr){19-21}
& & &
\textbf{Pre} & \textbf{Recall} & \textbf{F1} &
\textbf{Pre} & \textbf{Recall} & \textbf{F1} &
\textbf{Pre} & \textbf{Recall} & \textbf{F1} &
\textbf{Pre} & \textbf{Recall} & \textbf{F1} &
\textbf{Pre} & \textbf{Recall} & \textbf{F1} &
\textbf{Pre} & \textbf{Recall} & \textbf{F1} \\
\midrule

\multirow{4}{*}{\rotatebox[origin=c]{90}{Zero-shot}}
& Gemini-2.5~\cite{comanici2025gemini} & -
& 42.86 & 42.66 & 42.76
& 55.00 & 47.50 & 50.98
& 54.55 & 56.55 & 55.53
& 42.86 & 58.00 & 50.30
& 19.99 & 47.35 & 14.99
& 43.05 & 50.41 & 42.91 \\

& Qwen2.5-VL-72B~\cite{bai2025qwen2} & -
& 44.81 & 49.87 & 47.20
& 36.08 & 40.38 & 38.11
& 45.61 & 46.50 & 46.05
& 41.54 & 44.91 & 43.16
& 47.00 & 42.64 & 44.71
& 43.01 & 44.86 & 43.85 \\

& DeepSeek-VL2-27B~\cite{wu2024deepseekvl2mixtureofexpertsvisionlanguagemodels} & -
& 41.21 & 46.18 & 43.55
& 46.67 & 43.18 & 44.86
& 48.11 & 50.12 & 49.09
& 38.95 & 37.01 & 37.95
& 41.29 & 47.56 & 44.20
& 43.24 & 44.81 & 43.93 \\

& GPT-4o~\cite{achiam2023gpt} & -
& 60.00 & 59.24 & 60.59
& 43.70 & 37.58 & 40.41
& 41.34 & 45.22 & 42.78
& 40.51 & 51.59 & 45.36
& 44.58 & 31.21 & 35.32
& 46.03 & 44.97 & 44.89 \\
\midrule

& FKA\_OWL~\cite{liu2024fka} & MM'24
& 78.71 & 93.47 & 85.40
& 88.18 & 72.08 & 78.32
& 65.37 & 56.09 & 60.38
& 49.91 & 21.67 & 30.22
& 61.35 & 69.97 & 65.38
& 68.70 & 62.65 & 64.15 \\

& HAMMER~\cite{shao2024detecting} & CVPR'24
& 82.98 & 80.93 & 81.94
& 74.98 & 73.67 & 74.32
& 71.21 & 73.86 & 72.51
& 52.12 & 51.30 & 51.70
& \textbf{74.17} & 72.97 & 73.57
& 71.09 & 70.55 & 70.81 \\

& MIMOE~\cite{liu2025modality_interactive} & WWW'25
& 81.33 & 86.51 & 83.84
& 80.06 & 66.31 & 72.54
& 69.12 & \textbf{78.02} & \textbf{73.30}
& 56.81 & 51.45 & 53.45
& 73.56 & 65.86 & 69.50
& 72.58 & 69.83 & 70.93 \\

& GLPN~\cite{hu2025synergizing} & ACL'25
& 59.81 & \textbf{95.45} & 73.54
& 69.80 & 57.79 & 63.23
& 51.15 & 12.66 & 20.30
& 22.22 & 0.81 & 1.57
& 54.75 & 27.60 & 36.70
& 51.55 & 38.86 & 39.07 \\

\rowcolor{tabgroup}
\multirow{-5}{*}{\rotatebox[origin=c]{90}{OmniFake}}
& \textbf{UMFDet(Ours)} & -
& \textbf{87.04} & 92.94 & \textbf{89.89}
& \textbf{92.17} & \textbf{82.22} & \textbf{86.92}
& \textbf{79.49} & 63.23 & 70.43
& \textbf{71.95} & \textbf{61.85} & \textbf{66.52}
& 68.68 & \textbf{81.17} & \textbf{74.40}
& \textbf{82.56} & \textbf{82.53} & \textbf{82.23} \\

\bottomrule
\end{tabular}%
}
\vspace{-0.3cm}
\end{table*}

\begin{table*}[t]
\centering
\caption{Comparison of multimodal learning methods on DGM4. The best results in each group are highlighted in bold. ``AVG'' denotes performance over the entire dataset.}
\label{tab:dgm4_comparison}
\vspace{-0.3cm}
\small
\setlength{\tabcolsep}{4.3pt}
\renewcommand{\arraystretch}{1.08}

\begin{tabular}{llcccccccccccc}
\toprule
\multirow{2}{*}{\textbf{Method}} &
\multirow{2}{*}{\textbf{Venue}} &
\multicolumn{4}{c}{\textbf{AVG}} &
\multicolumn{4}{c}{\textbf{Real}} &
\multicolumn{4}{c}{\textbf{AI-synthesized}} \\
\cmidrule(lr){3-6}
\cmidrule(lr){7-10}
\cmidrule(lr){11-14}

& & \textbf{ACC} & \textbf{Pre} & \textbf{Recall} & \textbf{F1}
& \textbf{ACC} & \textbf{Pre} & \textbf{Recall} & \textbf{F1}
& \textbf{ACC} & \textbf{Pre} & \textbf{Recall} & \textbf{F1} \\
\midrule

Qwen-2.5-VL-72B~\cite{bai2025qwen2} & -
& 60.06 & 57.48 & 57.81 & 57.64
& 60.46 & 42.71 & 50.92 & 46.46
& 60.12 & 72.26 & 64.71 & 68.28 \\

GPT-4o~\cite{achiam2023gpt} & -
& 62.14 & 59.20 & 58.45 & 58.82
& 62.75 & 45.03 & 52.86 & 48.56
& 61.93 & 74.10 & 66.48 & 69.55 \\

DeepSeek-VL2-27B~\cite{wu2024deepseekvl2mixtureofexpertsvisionlanguagemodels}
& -
& 57.23 & 62.51 & 57.12 & 59.69
& 59.54 & \textbf{57.17} & \textbf{66.14} & \textbf{61.33}
& 43.21 & 67.82 & 48.03 & 56.23 \\

Gemini-2.5~\cite{comanici2025gemini} & -
& \textbf{67.20} & \textbf{64.05} & \textbf{63.30} & \textbf{63.67}
& \textbf{67.90} & 48.60 & 57.10 & 52.51
& \textbf{66.85} & \textbf{78.90} & \textbf{71.20} & \textbf{74.85} \\
\midrule

MIMOE~\cite{liu2025modality_interactive} & WWW'25
& 62.11 & 58.74 & 59.32 & 58.84
& 62.11 & 44.02 & 50.98 & 47.25
& 62.11 & 73.45 & 67.65 & 70.43 \\

FKA-OWL~\cite{liu2024fka} & MM'24
& 76.37 & 76.91 & 78.58 & 79.30
& 69.62 & \textbf{74.09} & 70.47 & \textbf{82.82}
& 76.14 & \textbf{93.92} & 72.67 & 81.94 \\

HAMMER~\cite{shao2024detecting} & CVPR'24
& 80.62 & 81.27 & \textbf{85.13} & 83.16
& 80.63 & 66.87 & 82.83 & 74.00
& 80.63 & 84.28 & \textbf{89.53} & \textbf{86.83} \\

\rowcolor{tabgroup}
\textbf{UMFDet (Ours)} & -
& \textbf{83.10} & \textbf{81.35} & 84.62 & \textbf{83.20}
& \textbf{86.88} & 69.21 & \textbf{86.88} & 77.14
& \textbf{81.18} & 92.85 & 81.18 & 86.62 \\

\bottomrule
\end{tabular}
\vspace{-0.2cm}
\end{table*}

\begin{figure*}[t]
\centering

\begin{subfigure}[t]{0.32\textwidth}
    \centering
    \includegraphics[width=\linewidth]
    {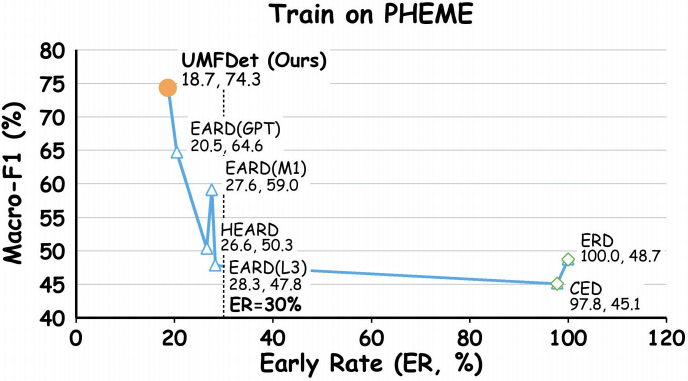}
    \caption{Early profile.}
    \label{fig:early_profile}
\end{subfigure}
\hfill
\begin{subfigure}[t]{0.32\textwidth}
    \centering
    \includegraphics[width=\linewidth]
    {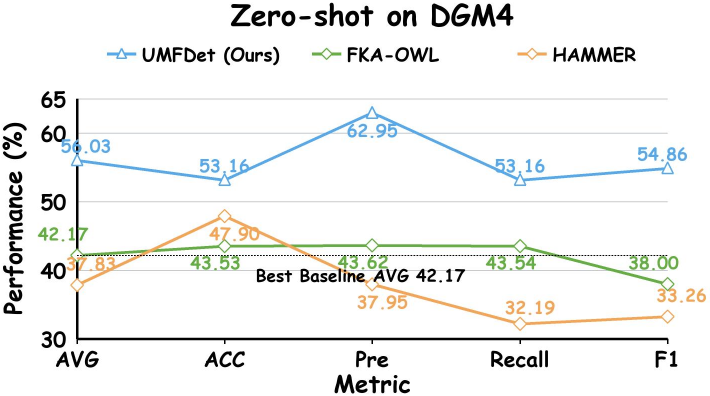}
    \caption{DGM4 profile.}
    \label{fig:dgm4_profile}
\end{subfigure}
\hfill
\begin{subfigure}[t]{0.32\textwidth}
    \centering
    \includegraphics[width=\linewidth]
    {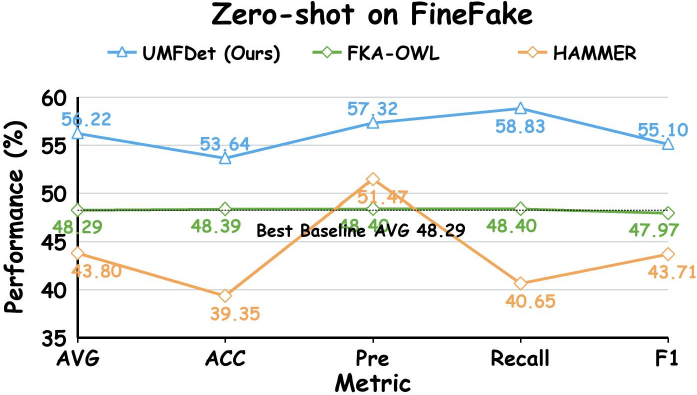}
    \caption{FineFake profile.}
    \label{fig:finefake_profile}
\end{subfigure}
\vspace{-0.3cm}
\caption{Performance profiles of UMFDet under different evaluation settings.}
\label{fig:umfdet_profiles}
\vspace{-0.3cm}
\end{figure*}

Finally, the total training objective of our UMFDet is a joint optimization of the language modeling loss $\mathcal{L}_{Det}$, the expert-wise discriminative regularization $\mathcal{L}_{EDR}$, and the alignment loss $\mathcal{L}_{CCA}$:
\begin{equation}
    \mathcal{L} = \mathcal{L}_{Det} + \lambda (\mathcal{L}_{EDR} + \mathcal{L}_{CCA}).
\end{equation}

\begin{table*}[t]
\centering
\caption{Zero-shot binary detection results (\%) on the MMFakeBench validation and test sets. We compare UMFDet with the ``Standard'' and ``MMD-Agent'' prompting baselines defined in MMFakeBench.}
\label{tab:mmfakebench_binary}
\vspace{-0.3cm}
\small
\setlength{\tabcolsep}{4.3pt}
\renewcommand{\arraystretch}{1.08}

\begin{tabular}{lllcccccccc}
\toprule
\textbf{Model} & \textbf{Language} & \textbf{Prompt} &
\multicolumn{4}{c}{\textbf{Validation (1000)}} &
\multicolumn{4}{c}{\textbf{Test (10000)}} \\
\cmidrule(lr){4-7}
\cmidrule(lr){8-11}

\textbf{Name} & \textbf{Model} & \textbf{Method} &
\textbf{F1} & \textbf{Precision} & \textbf{Recall} & \textbf{ACC} &
\textbf{F1} & \textbf{Precision} & \textbf{Recall} & \textbf{ACC} \\
\midrule

\rowcolor{tabgroup}
\multicolumn{11}{c}{\textbf{LVLMs with 7B Parameters}} \\

InstructBLIP~\cite{dai2023instructblip}
& Vicuna-7B~\cite{dai2023instructblip}
& Standard
& 14.7 & 30.8 & 13.2 & 8.1
& 16.1 & 40.5 & 14.2 & 8.8 \\

Qwen-VL~\cite{bai2025qwen2}
& Qwen-7B~\cite{bai2025qwen2}
& Standard
& 43.6 & 50.6 & 44.9 & 60.3
& 44.0 & 51.6 & 45.2 & 60.5 \\

PandaGPT~\cite{su2023pandagpt}
& Vicuna-7B~\cite{dai2023instructblip}
& Standard
& 24.6 & 60.6 & 50.5 & 30.9
& 24.1 & 61.7 & 50.4 & 30.6 \\

mPLUG-Owl2~\cite{ye2024mplugowl3longimagesequenceunderstanding}
& LLaMA2-7B~\cite{li2024llavanext-ablations}
& Standard
& 47.2 & 64.9 & 52.3 & 70.6
& 48.7 & 71.1 & 53.3 & 71.4 \\

LLaVA-1.6~\cite{li2024llavanext-ablations}
& Vicuna-7B~\cite{dai2023instructblip}
& Standard
& 48.1 & 48.2 & 48.5 & 59.5
& 52.5 & 53.0 & 52.6 & 62.5 \\

\rowcolor{tabgroup}
\multicolumn{11}{c}{\textbf{LVLMs with 13B Parameters}} \\

InstructBLIP~\cite{dai2023instructblip}
& Vicuna-13B~\cite{dai2023instructblip}
& Standard
& 41.1 & 35.0 & 49.9 & 69.9
& 41.1 & 35.0 & 49.9 & 69.8 \\

& & MMD-Agent
& 51.3 & 53.4 & 54.0 & 53.1
& 47.9 & 50.1 & 50.1 & 49.9 \\

LLaVA-1.6~\cite{li2024llavanext-ablations}
& Vicuna-13B~\cite{dai2023instructblip}
& Standard
& 41.1 & 35.0 & 50.0 & 69.7
& 42.3 & 57.3 & 50.1 & 69.5 \\

& & MMD-Agent
& 51.8 & 66.7 & 54.6 & \textbf{71.4}
& 50.2 & 67.3 & 53.9 & 71.3 \\

\rowcolor{tabgroup}
\multicolumn{11}{c}{\textbf{Train on OmniFake}} \\

HAMMER~\cite{shao2024detecting}
& -
& -
& 56.68 & 69.54 & 57.62 & 57.64
& 57.29 & \textbf{71.32} & 58.11 & 58.21 \\

\rowcolor{tabgroup}
\textbf{UMFDet(Ours)}
& Florence-2-0.7B~\cite{xiao2024florence}
& Ours (Sec.~3.2)
& \textbf{61.74} & \textbf{70.16}
& \textbf{61.14} & \textbf{61.17}
& \textbf{61.70} & 70.94
& \textbf{61.12} & \textbf{61.14} \\

\bottomrule
\end{tabular}
\vspace{-0.2cm}
\end{table*}

\begin{table*}[t]
\centering
\caption{Ablation study including (a) component-wise analysis, (b) number of MoE layers, (c) discussion of hyper-parameter $\lambda$, (d) larger backbone validation, (e) different numbers of experts, and (f) open-world generalization.}
\label{tab:ablation_study}
\vspace{-0.2cm}
\scriptsize
\renewcommand{\arraystretch}{1.05}

\begin{minipage}[t]{0.36\textwidth}
\centering
\vspace{0pt}
\setlength{\tabcolsep}{3.8pt}

\begin{tabular}{ccc|cccc}
\rowcolor{tabhead}
\multicolumn{3}{c|}{\textbf{Components}}
& \textbf{ACC} & \textbf{Prec.} & \textbf{Rec.} & \textbf{F1} \\
\rowcolor{tabhead}
\textbf{CMoE} & \textbf{EDR} & \textbf{CCA}
& & & & \\
\midrule[1.2pt]

\cmark & &
& 76.35 & 76.71 & 76.35 & 75.99 \\

\cmark & \cmark &
& 79.71 & 80.33 & 76.58 & 78.03 \\

\rowcolor{tabgroup}
\cmark & \cmark & \cmark
& \textbf{82.53} & 82.56 & \textbf{82.53} & 82.23 \\
\end{tabular}

\vspace{2pt}
\textbf{(a)} Component-wise analysis
\end{minipage}
\hfill
\begin{minipage}[t]{0.29\textwidth}
\centering
\vspace{0pt}
\setlength{\tabcolsep}{4.5pt}

\begin{tabular}{c|cccc}
\rowcolor{tabhead}
\textbf{MoE} & \multicolumn{4}{c}{\textbf{Metrics}} \\
\rowcolor{tabhead}
\textbf{Layers} & \textbf{ACC} & \textbf{Prec.}
& \textbf{Rec.} & \textbf{F1} \\
\midrule[1.2pt]

0 & 74.12 & 76.05 & 74.12 & 74.63 \\

\rowcolor{tabgroup}
1 & \textbf{82.53} & 82.56 & \textbf{82.53} & 82.23 \\

2 & 80.31 & 80.62 & 79.51 & 80.06 \\
\end{tabular}

\vspace{2pt}
\textbf{(b)} Number of MoE layers
\end{minipage}
\hfill
\begin{minipage}[t]{0.29\textwidth}
\centering
\vspace{0pt}
\setlength{\tabcolsep}{4.5pt}

\begin{tabular}{c|cccc}
\rowcolor{tabhead}
\textbf{Weight} & \multicolumn{4}{c}{\textbf{Metrics}} \\
\rowcolor{tabhead}
\textbf{$\lambda$} & \textbf{ACC} & \textbf{Prec.}
& \textbf{Rec.} & \textbf{F1} \\
\midrule[1.2pt]

0.01 & 79.26 & 76.92 & 81.33 & 80.12 \\

\rowcolor{tabgroup}
0.05 & \textbf{82.53} & 82.56 & \textbf{82.53} & 82.23 \\

0.1 & 77.20 & 76.97 & 77.20 & 76.45 \\
\end{tabular}

\vspace{2pt}
\textbf{(c)} Hyper-parameter $\lambda$
\end{minipage}

\vspace{7pt}

\begin{minipage}[t]{0.36\textwidth}
\centering
\vspace{0pt}
\setlength{\tabcolsep}{2.9pt}

\begin{tabular}{l|ccc}
\rowcolor{tabhead}
& \multicolumn{3}{c}{\textbf{Metrics}} \\
\rowcolor{tabhead}
\multirow{-2}{*}{\textbf{Method \& Backbone}}
& \textbf{Pre.} & \textbf{Rec.} & \textbf{F1} \\
\midrule[1.2pt]

Qwen2.5-VL-3B (Stand.)
& 88.20 & 77.13 & 79.34 \\

UMFDet (Qwen2.5-VL-3B)
& \textbf{88.41} & 81.25 & \textbf{82.88} \\

\rowcolor{tabgroup}
\textbf{UMFDet (Florence-2-0.7B)}
& 82.56 & \textbf{82.53} & 82.23 \\
\end{tabular}

\vspace{10pt}
\textbf{(d)} Larger backbone validation
\end{minipage}
\hfill
\begin{minipage}[t]{0.29\textwidth}
\centering
\vspace{0pt}
\setlength{\tabcolsep}{5.0pt}

\begin{tabular}{c|ccc}
\rowcolor{tabhead}
& \multicolumn{3}{c}{\textbf{Metrics}} \\
\rowcolor{tabhead}
\multirow{-2}{*}{\textbf{Experts}}
& \textbf{Pre.} & \textbf{Rec.} & \textbf{F1} \\
\midrule[1.2pt]

2 & 76.64 & 76.59 & 76.56 \\

\rowcolor{tabgroup}
\textbf{3 (Ours)}
& \textbf{82.56} & \textbf{82.53} & \textbf{82.23} \\

4 & 79.63 & 80.21 & 79.92 \\

5 & 80.08 & 75.76  & 77.39 \\
\end{tabular}

\vspace{2pt}
\textbf{(e)} Number of experts
\end{minipage}
\hfill
\begin{minipage}[t]{0.29\textwidth}
\centering
\vspace{0pt}
\setlength{\tabcolsep}{4.5pt}

\begin{tabular}{l|ccc}
\rowcolor{tabhead}
& \multicolumn{3}{c}{\textbf{Methods}} \\
\rowcolor{tabhead}
\multirow{-2}{*}{\textbf{Metric}}
& \textbf{Ours} & \textbf{HAMMER} & \textbf{FKA-OWL} \\
\midrule[1.2pt]

ACC & \textbf{52.00} & 48.00 & 47.00 \\
F1  & \textbf{51.04} & 46.26 & 41.37 \\
\end{tabular}

\vspace{16pt}
\textbf{(f)} Open-world test
\end{minipage}
\vspace{-0.3cm}
\end{table*}

\section{Experiment}
\textbf{Experiment Setup.} 
Experimental details and more quantitative and qualitative results are provided in the Supplementary Material.

\subsection{Comparison with State-of-the-Art Methods}

\noindent\textbf{Quantitative Results.} To evaluate the performance of the proposed framework, we conduct a comprehensive comparison with a wide range of state-of-the-art methods on the OmniFake dataset. As summarized in Table~\ref{tab:omnifake_main}, UMFDet substantially outperforms all baseline models under the unified five-class classification setting. Specifically, it achieves an average precision of 82.56\%, an average recall of 82.53\%, and  F1-score of 82.23\% across the five categories. 

UMFDet performs strongly across challenging subsets. It achieves an F1-score of 89.89\% on Real, surpassing the best baseline by 4.49\%. On Human-crafted, it obtains 92.17\% precision, 82.22\% recall, and 86.92\% F1-score. For Text Manipulation, UMFDet reaches 71.95\% precision, 61.85\% recall, and 66.52\% F1-score, improving the strongest baseline by 13.07\%. It also achieves the best recall and F1-score on Mixed Manipulation, at 81.17\% and 74.40\%. These results demonstrate its robustness across diverse misinformation types.

\noindent\textbf{Performance on MMFakeBench.} 
We evaluate UMFDet on the mixed-source MMFakeBench dataset~\cite{liu2024mmfakebench}. As shown in Table~\ref{tab:mmfakebench_binary}, it achieves the best results on both splits, with 61.74\% F1 and 61.17\% accuracy on validation, and 61.70\% F1 and 61.14\% accuracy on test. It also surpasses the best Standard 7B model (52.5\% F1) and MMD-Agent-enhanced 13B model (50.2\% F1), demonstrating robust mixed-source misinformation detection.

\noindent\textbf{Performance on DGM4.} 
We further evaluate UMFDet on DGM4~\cite{shao2024detecting}. As shown in Table~\ref{tab:dgm4_comparison}, it achieves the best overall accuracy and F1-score of 83.10\% and 83.20\%, respectively. On the Real subset, UMFDet obtains the highest accuracy and recall, both at 86.88\%, surpassing HAMMER by 6.25\% and 4.05\%. For AI-synthesized samples, it achieves the best precision of 92.85\%, exceeding HAMMER by 8.57\%.

\noindent\textbf{Performance on PHEME.}
As shown in Figure~\ref{fig:early_profile}, UMFDet achieves 74.3\% F1-score, surpassing the best LLM-based variant, EARD w/ ChatGPT, by 9.7\%. It also obtains the lowest Early Rate of 18.7\%, improving over HEARD by 7.9\%.

\noindent\textbf{Generalization Evaluation.}
UMFDet achieves the best zero-shot performance across all metrics on DGM4, with an F1-score of 54.86\%, exceeding FKA-OWL by 16.86\% (Figure~\ref{fig:dgm4_profile}). On FineFake~\cite{zhou2024finefake}, it obtains 53.64\% accuracy and 55.10\% F1, outperforming FKA-OWL and HAMMER by 7.13\% and 11.39\%, respectively (Figure~\ref{fig:finefake_profile}). These results demonstrate strong cross-domain generalization.
\vspace{-0.3cm}
\subsection{Ablation Study}

\noindent\textbf{Contributions of CMoE, EDR, and CCA.}
As shown in Table~\ref{tab:ablation_study}(a), CMoE achieves 76.35\% ACC and 75.99\% F1. Adding EDR improves them to 79.71\% and 78.03\%, while CCA further yields the best results of 82.53\% ACC, 82.56\% precision, 82.53\% recall, and 82.23\% F1, confirming the complementary benefits of expert specialization, discriminative regularization, and cross-modal alignment.

\noindent\textbf{Effect of the Number of MoE Layers.}
Table~\ref{tab:ablation_study}(b) shows that one CMoE layer performs best, achieving 82.53\% ACC and 82.23\% F1, compared with 74.12\% and 74.63\% without CMoE. Using two layers reduces them to 80.31\% and 80.06\%, indicating that deeper routing introduces unnecessary complexity. We therefore use one layer.

\noindent\textbf{Effect of the Hyper-parameter $\lambda$.}
As reported in Table~\ref{tab:ablation_study}(c), $\lambda=0.05$ achieves the best ACC and F1 of 82.53\% and 82.23\%. Smaller or larger values degrade performance, with 79.26\%/80.12\% at $\lambda=0.01$ and 77.20\%/76.45\% at $\lambda=0.1$. We thus set $\lambda=0.05$.

\noindent\textbf{Performance with a Larger Backbone.}
As shown in Table~\ref{tab:ablation_study}(d), UMFDet with Qwen2.5-VL-3B improves precision and F1 from 88.20\% and 79.34\% to 88.41\% and 82.88\%, respectively. We use Florence-2-0.7B as the default backbone for a better effectiveness--efficiency trade-off.

\noindent\textbf{Number of Experts.}
Table~\ref{tab:ablation_study}(e) shows that three experts achieve the best overall performance, reaching 82.56\% precision, 82.53\% recall, and 82.23\% F1. Both two- and four-expert configurations perform worse, supporting our default design.

\noindent\textbf{Open-world Test.}
We additionally collect 100 real-time news samples for evaluation. As shown in Table~\ref{tab:ablation_study}(f), UMFDet achieves the best accuracy (52.00\%) and F1-score (51.04\%), outperforming HAMMER and FKA-OWL in this challenging open-world setting.


\vspace{0.8cm}
\section{Conclusion}

In this work, we tackle multimodal misinformation spanning both human-crafted and AI-generated manipulations. We introduce OmniFake, a 98K-sample benchmark covering five categories: \emph{Real}, \emph{Human-crafted}, \emph{Vision Manipulation}, \emph{Text Manipulation}, and \emph{Mixed Manipulation}. We propose UMFDet, a unified detection framework that combines a VLM backbone, a Category-aware MoE adapter, an expert-wise discriminative regularization mechanism, and cross-modal consistency alignment to enhance the experts' capability of modeling category-specific cues and image--text semantic correlations. Extensive experiments demonstrate that UMFDet achieves robust and generalizable performance, surpassing specialized baselines and providing a practical solution for unified multimodal misinformation detection.
\clearpage
\bibliographystyle{ACM-Reference-Format}
\bibliography{sample-base}

\clearpage
\appendix

\section*{Supplementary Material}

\noindent\textbf{Contents}\par
\vspace{2pt}
\begin{tabular*}{\columnwidth}{@{\extracolsep{\fill}}llr@{}}
\hyperref[app:related]{A}   &
\hyperref[app:related]{Related Work} &
\pageref{app:related} \\

\hyperref[app:prompt]{B}    &
\hyperref[app:prompt]{Prompt Paradigm of UMFDet} &
\pageref{app:prompt} \\

\hyperref[app:datasets]{C}  &
\hyperref[app:datasets]{Details on Datasets} &
\pageref{app:datasets} \\

\hyperref[app:baselines]{D} &
\hyperref[app:baselines]{Baselines} &
\pageref{app:baselines} \\

\hyperref[app:setup]{E}     &
\hyperref[app:setup]{Experiment Setup} &
\pageref{app:setup} \\

\hyperref[app:setup]{F}     &
\hyperref[app:setup]{Expert Routing and Error Analysis} &
\pageref{app:error} \\

\hyperref[app:setup]{G}     &
\hyperref[app:setup]{Dataset examples} &
\pageref{app:exp} \\

\hyperref[app:setup]{H}     &
\hyperref[app:setup]{Representative construction examples of OmniFake} &
\pageref{app:show} \\

\end{tabular*}
\vspace{6pt}

\section{Related Work}
\label{app:related}
\subsection{Fake News Detection}

With the rapid advancement of generative models and Large Language Models (LLMs), synthetic content on social media has proliferated, raising concerns about information credibility and platform governance. Existing fake news detection methods can be broadly divided into unimodal and multimodal paradigms. Unimodal approaches process either images or text independently~\cite{ghai2024deep,xiao2024msynfd,10.1145/3690624.3709406, 10.1145/3637528.3671977, 10.1145/3580305.3599873}, limiting their ability to capture cross-modal deceptive cues. More recently, multimodal detection has attracted increasing attention~\cite{peng2024not,luvembe2024caf}, with most methods relying on cross-modal contrastive learning or feature fusion to enhance vision--language representations and semantic alignment. Representative methods include COOLANT~\cite{wang2023cross}, which employs an auxiliary-task-enhanced contrastive framework with softened negatives and attention-based aggregation, and FKA-OWL~\cite{liu2024fka}, which leverages world knowledge from large vision--language models to improve cross-domain generalization. Despite this progress, most existing systems remain restricted to binary classification and provide limited support for the multi-class and multi-source manipulation scenarios commonly encountered in the wild, thereby constraining their applicability.

\subsection{MultiModal Large Language Models}
\begin{sloppypar}

Research in this field has evolved along a discernible trajectory. 
CLIP \cite{radford2021learning} and ALIGN \cite{li2021align} established robust cross-modal alignment and zero-shot recognition via contrastive learning, substantially improving image-text matching and retrieval. 
SimVLM \cite{wang2021simvlm} then advanced language-centric, caption-driven pretraining to strengthen generation and transfer. 
To reduce the cost of end-to-end scaling, a decoupled paradigm emerged: BLIP-2 \cite{li2023blip} and CoCa \cite{yu2022coca} connect a vision encoder to a language model through cross-attention or lightweight bridges. 
Building on this, multimodal instruction alignment became central, with InstructBLIP \cite{dai2023instructblip} and Qwen2-VL \cite{wang2024qwen2} demonstrating strong improvements in open-ended QA and reasoning via instruction tuning. 
More recently, Florence-2 \cite{xiao2024florence} epitomizes the trend of using language as a unifying interface to aggregate large-scale priors and transfer them to downstream tasks.

\end{sloppypar}

\section{Prompt paradigm of UMFDet}
\label{app:prompt}
The details of the  question-answer prompts are as follows:

``\emph{Task: Multimodal Fake News Classification. 
The news title is: \texttt{\{news title\}}. 
Analyze the provided image and news title. Categorize the content into exactly one of the following five categories:}

\emph{\textbf{real}: Authentic news. Both the image and text are entirely untampered and human-created.}

\emph{\textbf{human-crafted}: The image is authentic, but the title contains misleading information, false claims, or out-of-context text.}

\emph{\textbf{vision manipulation}: The image is manipulated while the text remains authentic. Visual manipulation explicitly includes AI-generated images, image inpainting, face swapping, or face attribute editing.}

\emph{\textbf{text manipulation}: The image is authentic, but the text has been altered, e.g., keyword replaced or AI-generated.}

\emph{\textbf{mixed manipulation}: Both the image and the text have been manipulated, e.g., the image contains AI generation, inpainting, face swapping, or attribute editing, and the text is also altered.}

\emph{The news belongs to:}

\section{Details on Datasets.} 
\label{app:datasets}
    
To evaluate the generalizability and robustness of UMFDet, we conduct extensive experiments across four diverse benchmarks. These datasets encompass a wide range of misinformation types, from visual manipulation to linguistic deception, ensuring a comprehensive assessment of our method.

\noindent\textbf{MMFakeBench:} MMFakeBench ~\cite{liu2024mmfakebench} focuses on the emerging challenge of mixed-source misinformation. It collaborates various generative models and AI tools to create sophisticated distortions, including textual veracity, visual veracity, and cross-modal consistency distortions.

\noindent\textbf{DGM4:} DGM4 ~\cite{shao2024detecting} specifically targets fine-grained visual manipulation. Unlike traditional fake news datasets, DGM4 emphasizes structural and attribute-level tampering, particularly focusing on face-related image manipulations (e.g., face swapping, attribute editing) and their corresponding textual inconsistencies.

\noindent\textbf{PHEME:} PHEME ~\cite{zubiaga2016learning} is a representative dataset for rumor detection in breaking news scenarios. It consists of human-crafted textual rumors collected from social media (Twitter). 

\noindent\textbf{FineFake:} FineFake~\cite{zhou2024finefake} is a multi-domain, knowledge-enhanced benchmark covering topics and platforms. It focuses on human-fabricated news requiring external knowledge for verification.

\begin{figure*}[!t]
  \centering

  \makebox[\textwidth][c]{%
    \includegraphics[width=0.50\textwidth]{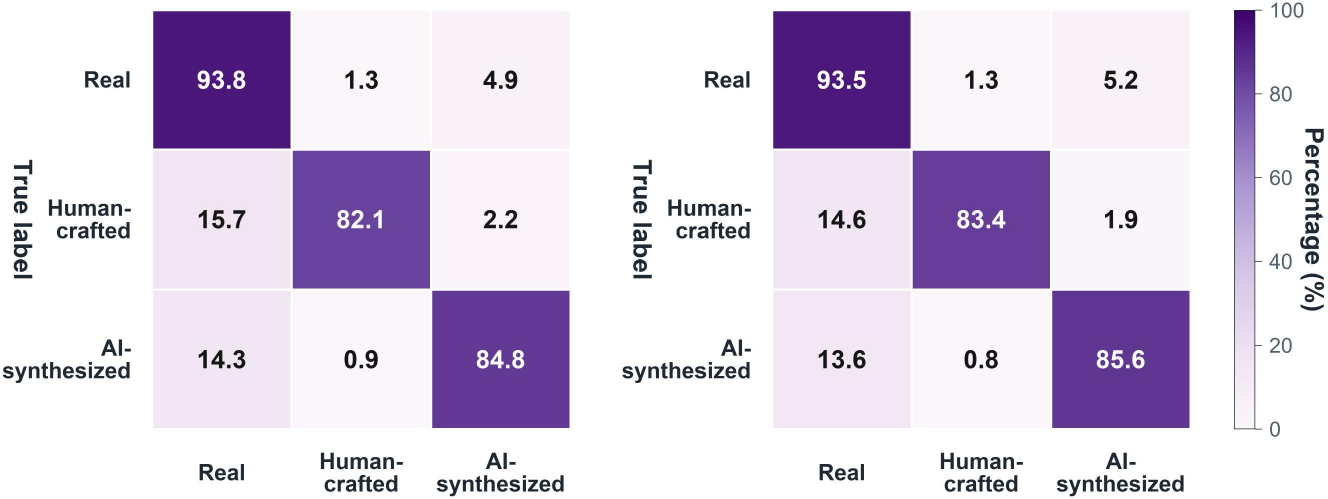}%
    \hspace{0.01\textwidth}%
    \includegraphics[width=0.47\textwidth]{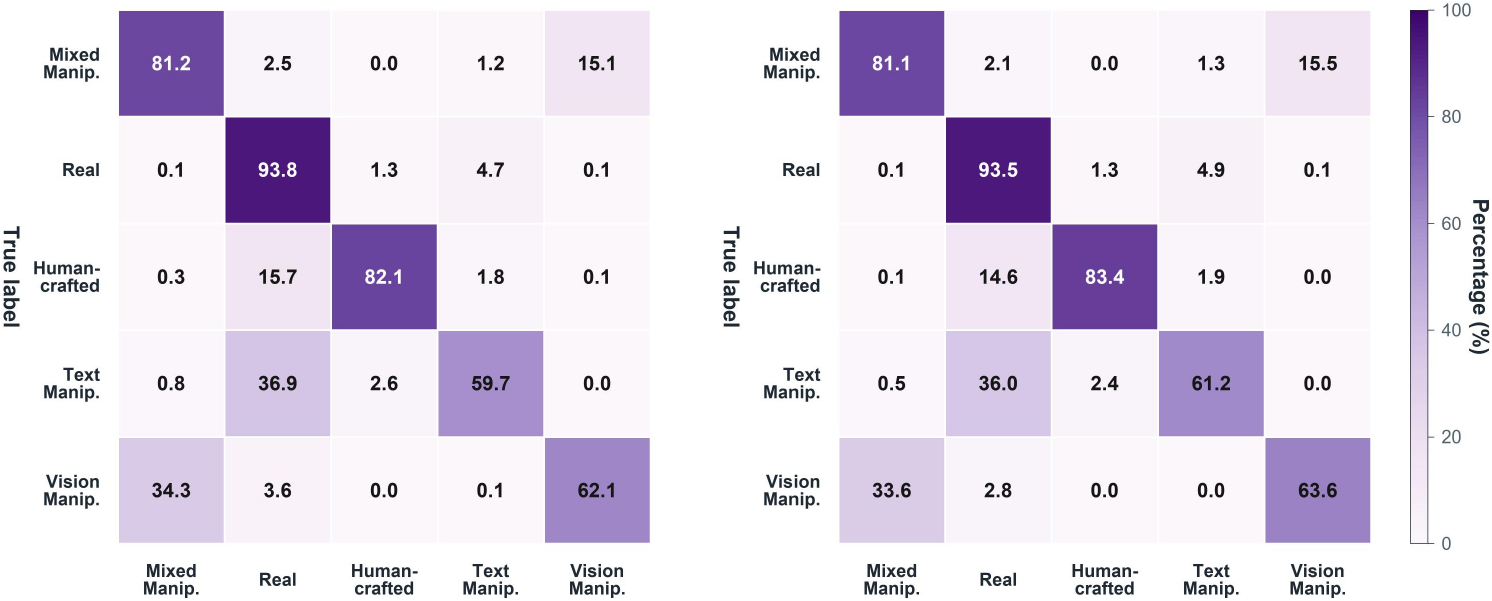}%
  }

  \caption{\textbf{Expert routing and error analysis.}
  Left: three-category routing distributions with CMoE and with EDR and CCA. Right: row-normalized five-class confusion matrices.}
  \label{fig:routing_error_analysis}
\end{figure*}

\begin{figure*}[!t]
  \centering
  \includegraphics[
    width=\textwidth,
    height=0.2\textheight
  ]{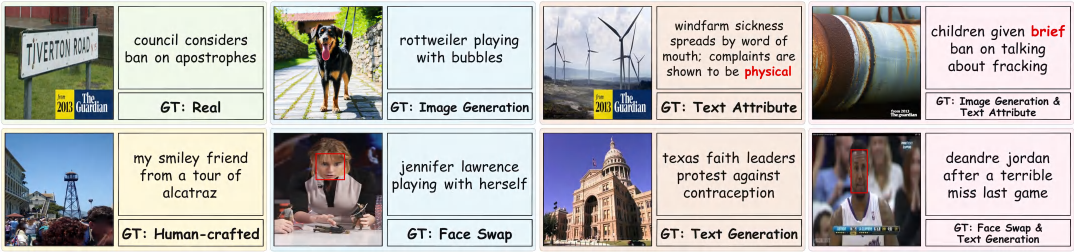}

  \caption{\textbf{Representative examples from OmniFake.}
  From left to right, the columns show \emph{Real}, \emph{Human-crafted},
  \emph{Vision Manipulation}, \emph{Text Manipulation}, and
  \emph{Mixed Manipulation} samples.}
  \label{fig:dataset_examples}
\end{figure*}

\section{Baselines.} 
\label{app:baselines}
To ensure a fair evaluation, we specifically adapt existing models that were originally designed for binary classification. 

\noindent\textbf{Inherent Generalizability of UMFDet:} Our proposed UMFDet demonstrates superior architectural flexibility. It can be seamlessly adapted to multi-class scenarios by simply adjusting the number of expert subspaces to match the required category count, requiring no structural modifications to the core alignment logic or feature extraction backbone.

\noindent\textbf{Structural Adaptation for Baselines:} In contrast, other competitive methods require specific re-configurations to handle the three-way classification task. Specifically, for HAMMER, GAME-ON, and COOLANT, we extend their final MLP-based classification heads or Softmax layers to produce three-way probability distributions. For the LVLM-based FKA-OWL, we modify both the output label space within the instruction prompts and the linear projection layers of its verification module. 

\section{Experiment Setup.} 
\label{app:setup}

We take Florence-2-0.7B~\cite{xiao2024florence} as the backbone. During training, we freeze the vision encoder and train the category-aware MoE module together with the language model’s encoder and decoder. This configuration balances performance and efficiency. All experiments are conducted on 4$\times$NVIDIA RTX 4090 GPUs. Our model produces the final category via parsing the class label from the \texttt{<answer>} field. We report Accuracy (ACC), and for fair comparison with prior methods, we also report Precision, Recall, and F1 score.

\noindent\textbf{Dataset Split.} We split OmniFake into training, validation, and testing sets with an 8:1:1 ratio while maintaining balanced category distributions across all splits. To prevent data leakage, we strictly ensure that samples derived from the same original source are assigned to only one split. Specifically, different manipulated versions of the same image--text pair, near-duplicate samples, and samples sharing identical source information are grouped together before splitting. This strategy guarantees that the evaluation sets contain unseen samples and provides a reliable assessment of model generalization.

\begin{table*}[!t]
\centering
\caption{Overall results (\%) on MMFakeBench validation and test sets. Standard refers to the default evaluation. MMD-Agent~\cite{liu2024mmfakebench} is a hierarchical framework.}
\label{tab:mmfakebench_overall}
\scriptsize
\setlength{\tabcolsep}{2.0pt}
\renewcommand{\arraystretch}{1.08}
\resizebox{\textwidth}{!}{
\begin{tabular}{l l l cccc cccc}
\toprule
\multirow{2}{*}{\textbf{Model}} &
\multirow{2}{*}{\textbf{Backbone}} &
\multirow{2}{*}{\textbf{Method}} &
\multicolumn{4}{c}{\textbf{Validation (1K)}} &
\multicolumn{4}{c}{\textbf{Test (10K)}} \\
\cmidrule(lr){4-7}
\cmidrule(lr){8-11}
& & &
\textbf{F1} & \textbf{Pre.} & \textbf{Rec.} & \textbf{Acc.} &
\textbf{F1} & \textbf{Pre.} & \textbf{Rec.} & \textbf{Acc.} \\
\midrule

Human & --- & ---
& 35.9 & 38.3 & 38.9 & 37.9
& -- & -- & -- & -- \\

\midrule
\rowcolor{tabgroup}
\multicolumn{11}{l}{\textit{LVLMs with 7B Parameter}} \\

Otter-Image~\cite{li2025otter} & MPT-7B & Stand.
& 5.2 & 10.5 & 3.4 & 4.1
& 4.9 & 9.3 & 3.3 & 4.0 \\

MiniGPT4~\cite{zhu2023minigpt} & Vicuna-7B & Stand.
& 5.2 & 5.2 & 21.2 & 9.0
& 5.3 & 6.9 & 21.0 & 9.1 \\

InstructBLIP~\cite{dai2023instructblip} & Vicuna-7B & Stand.
& 7.1 & 7.9 & 6.5 & 7.8
& 8.1 & 16.4 & 7.2 & 8.5 \\

Qwen-VL~\cite{wang2024qwen2} & Qwen-7B & Stand.
& 7.5 & 10.3 & 24.3 & 11.0
& 8.0 & 35.9 & 25.5 & 11.6 \\

VILA~\cite{lin2024vila} & LLaMA2-7B & Stand.
& 11.5 & 7.5 & 25.0 & 30.0
& 11.5 & 7.5 & 25.0 & 30.0 \\

PandaGPT~\cite{su2023pandagpt} & Vicuna-7B & Stand.
& 11.8 & 9.8 & 25.0 & 30.0
& 11.6 & 8.6 & 25.0 & 30.0 \\

mPLUG-Owl2~\cite{ye2024mplug} & LLaMA2-7B & Stand.
& 14.5 & 22.2 & 25.9 & 31.1
& 15.1 & 25.2 & 26.3 & 31.5 \\

BLIP2~\cite{li2023blip} & FlanT5-XL & Stand.
& 16.4 & 20.1 & 27.5 & 33.0
& 16.7 & 17.3 & 27.7 & 33.2 \\

LLaVA-1.6~\cite{li2024llavanext-ablations} & Vicuna-7B & Stand.
& 17.4 & 14.8 & 25.7 & 30.8
& 19.0 & 16.5 & 26.9 & 32.3 \\

\midrule
\rowcolor{tabgroup}
\multicolumn{11}{l}{\textit{LVLMs with 13B Parameter}} \\

\multirow{2}{*}{VILA~\cite{lin2024vila}} & \multirow{2}{*}{LLaMA2-13B} & Stand.
& 11.5 & 7.5 & 25.0 & 30.0
& 11.6 & 32.5 & 25.0 & 30.0 \\

& & Agent
& 22.7 & 27.3 & 24.4 & 28.7
& 24.0 & 30.4 & 25.5 & 29.4 \\

\multirow{2}{*}{InstructBLIP~\cite{dai2023instructblip}} & \multirow{2}{*}{Vicuna-13B} & Stand.
& 13.7 & 13.2 & 24.0 & 28.8
& 13.9 & 25.5 & 24.3 & 29.1 \\

& & Agent
& 26.0 & 33.3 & 30.1 & 29.5
& 24.5 & 32.1 & 28.8 & 27.3 \\

\multirow{2}{*}{BLIP2~\cite{li2023blip}} & \multirow{2}{*}{FlanT5-XXL} & Stand.
& 16.7 & 34.9 & 27.3 & 32.8
& 16.3 & 34.6 & 27.3 & 32.8 \\

& & Agent
& 31.6 & 39.8 & 32.2 & 34.4
& 28.8 & 39.0 & 30.4 & 32.1 \\

\multirow{2}{*}{LLaVA-1.6~\cite{li2024llavanext-ablations}} & \multirow{2}{*}{Vicuna-13B} & Stand.
& 12.0 & 22.5 & 25.0 & 30.0
& 14.4 & 35.7 & 26.0 & 31.2 \\

& & Agent
& 38.0 & 44.5 & 41.0 & 40.6
& 34.5 & 42.7 & 37.5 & 37.4 \\

\midrule
\rowcolor{tabgroup}
\multicolumn{11}{l}{\textit{LVLMs with 34B Parameter}} \\

LLaVA-1.6~\cite{li2024llavanext-ablations} & Nous-Hermes-2-Yi-34B & Stand.
& 25.7 & 44.5 & 33.7 & 40.4
& 25.4 & 44.1 & 33.8 & 40.5 \\

\midrule
\rowcolor{tabgroup}
\multicolumn{11}{c}{\textbf{Train on OmniFake}} \\

HAMMER~\cite{shao2024detecting} & - & -
& 56.68 & 69.54 & 57.62 & 57.64
& 57.29 & \textbf{71.32} & 58.11 & 58.21 \\

\rowcolor{tabgroup}
\textbf{Ours} & Florence-2-0.7B~\cite{xiao2024florence} & Ours (Sec.3.2)
& \textbf{61.74} & \textbf{70.16} & \textbf{61.14} & \textbf{61.17}
& \textbf{61.70} & 70.94 & \textbf{61.12} & \textbf{61.14} \\

\bottomrule
\end{tabular}
}
\end{table*}

\noindent\textbf{Evaluate on MMFakeBench.} We conduct zero-shot evaluation on MMFakeBench for the compared methods trained on the OmniFake dataset, to assess their generalization ability; the detailed results are shown in Table~\ref{tab:mmfakebench_overall}. The main findings are as follows.

Our UMFDet method achieves the highest F1 score on both splits and the best ACC
among the methods trained on OmniFake. In the test setting, UMFDet attains an F1
score of 61.70\% and an ACC of 61.14\%, outperforming HAMMER by 4.41\% in F1
and 2.93\% in ACC. It is also worth noting that our method substantially
outperforms the open-source LLaVA-1.6-34B model, whose F1 score under standard
prompting is 25.4\%. Under the MMD-Agent setting, UMFDet also surpasses
LLaVA-1.6-13B by 11.5\% in F1. In the validation setting, UMFDet again achieves
the highest F1 score of 61.74\%, together with an ACC of 61.17\%. Compared with
LLaVA-1.6-34B under standard prompting, our method gains 36.04\% in F1; compared
with LLaVA-1.6-13B with MMD-Agent, it improves F1 by 9.94\% while achieving
higher precision and recall. In particular, on both the validation and test
sets, our method substantially outperforms all 7B models and most 13B models in
F1. These results demonstrate the strong generalization capability of UMFDet
and further highlight the comprehensiveness of the OmniFake dataset.

\noindent\textbf{Zero-shot on DGM4.} We further evaluate the zero-shot cross-domain generalization of UMFDet on DGM4. As shown in Table~\ref{tab:zero_shot_dgm4}, UMFDet performs strongly across all test domains, including Guardian, USA Today, Washington Post, and BBC. It achieves an average accuracy of 53.16\% and an average F1-score of 54.86\%, surpassing the strongest baseline, FKA-OWL (38.00\% F1), by 16.86\%. On BBC, UMFDet obtains an F1-score of 60.96\%, outperforming FKA-OWL (38.41\%) by 22.55\%. These results demonstrate the strong robustness of UMFDet under unseen domain shifts.

\noindent\textbf{Zero-shot on FineFake.} We further evaluate the zero-shot generalization of UMFDet on FineFake. As shown in Table~\ref{tab:zeroshot_finefake_grouped}, UMFDet achieves the best overall performance, with 53.64\% accuracy and 55.10\% F1-score, surpassing FKA-OWL and HAMMER by 7.13 and 11.39 percentage points in F1, respectively. On the Human-crafted subset, UMFDet obtains the highest recall and F1-score of 71.65\% and 69.38\%, outperforming FKA-OWL by 20.47 percentage points in F1. These results demonstrate the strong zero-shot generalization of UMFDet to complex human-crafted misinformation.

\begin{table*}[t]
\centering
\caption{Zero-shot cross-domain results on DGM4 grouped by subsets. Best numbers per column are in bold. ``AVG'' denotes the average performance across the DGM4 news domains.}
\label{tab:zero_shot_dgm4}
\scriptsize
\setlength{\tabcolsep}{2.0pt}
\renewcommand{\arraystretch}{1.08}
\resizebox{\textwidth}{!}{
\begin{tabular}{l cccc ccc ccc ccc ccc}
\toprule
\multirow{2}{*}{\textbf{Method}} &
\multicolumn{4}{c}{\textbf{AVG}} &
\multicolumn{3}{c}{\textbf{Guardian}} &
\multicolumn{3}{c}{\textbf{USA Today}} &
\multicolumn{3}{c}{\textbf{Wash.\ Post}} &
\multicolumn{3}{c}{\textbf{BBC}} \\
\cmidrule(lr){2-5}
\cmidrule(lr){6-8}
\cmidrule(lr){9-11}
\cmidrule(lr){12-14}
\cmidrule(lr){15-17}
& \textbf{ACC} & \textbf{Pre} & \textbf{Recall} & \textbf{F1}
& \textbf{ACC} & \textbf{Pre} & \textbf{F1}
& \textbf{ACC} & \textbf{Pre} & \textbf{F1}
& \textbf{ACC} & \textbf{Pre} & \textbf{F1}
& \textbf{ACC} & \textbf{Pre} & \textbf{F1} \\
\midrule
HAMMER~\cite{shao2024detecting}
& 47.90 & 37.95 & 32.19 & 33.26
& 51.98 & 41.55 & 37.06
& \textbf{47.21} & 36.13 & 32.08
& \textbf{49.28} & 36.35 & 33.22
& 43.15 & 37.79 & 30.69 \\

FKA-OWL~\cite{liu2024fka}
& 43.53 & 43.62 & 43.54 & 38.00
& 37.81 & 38.64 & 35.52
& 46.75 & 46.37 & 38.70
& 47.04 & 46.76 & 39.37
& 42.52 & 42.69 & 38.41 \\

\rowcolor{tabgroup}
\textbf{UMFDet (Ours)}
& \textbf{53.16} & \textbf{62.95} & \textbf{53.16} & \textbf{54.86}
& \textbf{57.61} & \textbf{73.10} & \textbf{58.39}
& 46.58 & \textbf{58.11} & \textbf{49.22}
& 48.94 & \textbf{57.77} & \textbf{50.85}
& \textbf{59.50} & \textbf{62.80} & \textbf{60.96} \\
\bottomrule
\end{tabular}
}
\end{table*}

\begin{table*}[t]
\centering
\caption{Zero-shot results (\%) on \textit{FineFake}. ``AVG'' denotes the average performance over the Real and Human-crafted subsets.}
\label{tab:zeroshot_finefake_grouped}

\setlength{\tabcolsep}{3.8pt}
\renewcommand{\arraystretch}{1.10}

\resizebox{0.82\textwidth}{!}{%
\begin{tabular}{l cccc ccc ccc}
\toprule
\multirow{2}{*}{\textbf{Method}} &
\multicolumn{4}{c}{\textbf{AVG}} &
\multicolumn{3}{c}{\textbf{Real}} &
\multicolumn{3}{c}{\textbf{Human-crafted}} \\
\cmidrule(lr){2-5}
\cmidrule(lr){6-8}
\cmidrule(lr){9-11}
& \textbf{ACC} & \textbf{Pre} & \textbf{Recall} & \textbf{F1}
& \textbf{Pre} & \textbf{Recall} & \textbf{F1}
& \textbf{Pre} & \textbf{Recall} & \textbf{F1} \\
\midrule

HAMMER~\cite{shao2024detecting}
& 39.35 & 51.47 & 40.65 & 43.71
& 45.97 & \textbf{52.16} & \textbf{48.87}
& \textbf{56.96} & 29.13 & 38.55 \\

FKA-OWL~\cite{liu2024fka}
& 48.39 & 48.40 & 48.40 & 47.97
& 42.91 & 52.03 & 47.03
& 53.90 & 44.76 & 48.91 \\

\rowcolor{tabgroup}
\textbf{UMFDet (Ours)}
& \textbf{53.64} & \textbf{57.32} & \textbf{58.83} & \textbf{55.10}
& \textbf{58.82} & 46.00 & 40.81
& 55.81 & \textbf{71.65} & \textbf{69.38} \\

\bottomrule
\end{tabular}%
}
\end{table*}

\section{Expert Routing and Error Analysis}
\label{app:error}

As shown in Figure~\ref{fig:routing_error_analysis}, we analyze both expert routing behavior and five-class prediction errors. In the left panel, the first matrix presents the three-category routing distribution using CMoE, while the second shows the results after further incorporating EDR and CCA. The average routing accuracy improves from 90.43\% to 91.03\%, demonstrating that EDR and CCA facilitate more accurate expert specialization. Here, \emph{Real}, \emph{Human-crafted}, and \emph{AI-synthesized} denote the ground-truth categories and their corresponding routed expert groups.

The right panel presents row-normalized five-class confusion matrices for error analysis. Most errors occur between semantically related manipulation categories. In particular, real samples are occasionally confused with text-manipulated samples, text manipulation is frequently predicted as human-crafted misinformation, and vision manipulation is sometimes confused with mixed manipulation. These results highlight the difficulty of distinguishing fine-grained boundaries among \emph{Real}, \emph{Human-crafted}, \emph{Vision Manipulation}, \emph{Text Manipulation}, and \emph{Mixed Manipulation}.

\section{Dataset examples}
\label{app:exp}
Figure~\ref{fig:dataset_examples} presents representative image--text pairs from the five OmniFake categories. From left to right: (1) \textbf{Real}, where both the image and text are authentic; (2) \textbf{Human-crafted}, where an authentic image is paired with a human-written misleading headline; (3) \textbf{Vision Manipulation}, where the image is generated or edited while the text remains unchanged; (4) \textbf{Text Manipulation}, where the image is authentic but the headline is generated or semantically altered; and (5) \textbf{Mixed Manipulation}, where both modalities are manipulated. The examples cover diverse operations, including full-image generation, face manipulation, text rewriting, and their cross-modal combinations. Red bounding boxes indicate manipulated facial regions when applicable. Based on these patterns, OmniFake contains 98,592 samples and supports unified five-class multimodal misinformation detection.

\begin{figure*}[t]
  \centering
  \includegraphics[
    width=\textwidth,
    height=0.95\textheight
  ]{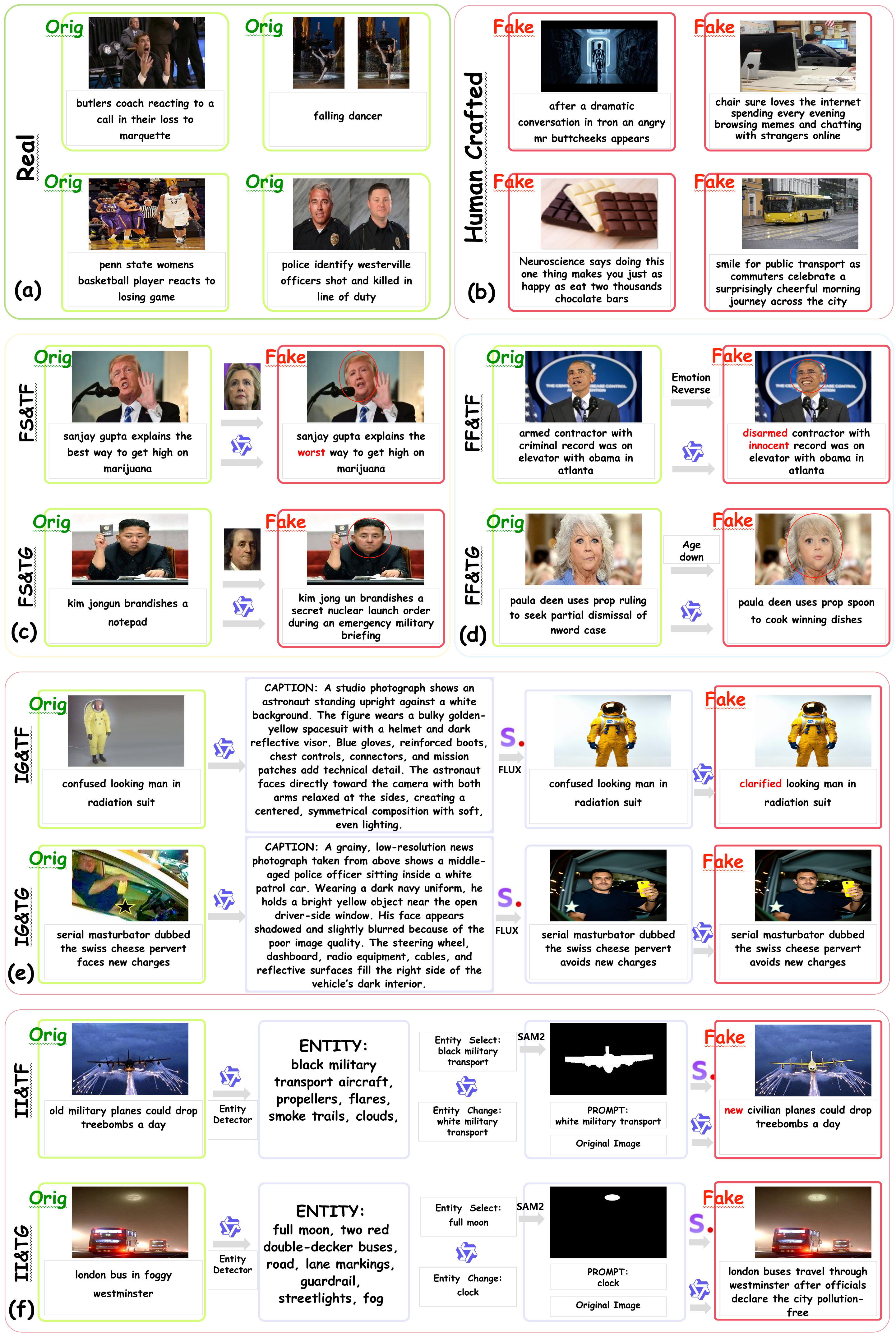}
  \caption{\textbf{Examples of multimodal manipulation construction in OmniFake.} Representative samples cover human-crafted misinformation, vision manipulation, text manipulation, and mixed manipulation scenarios.}
  \label{fig:dataset_show}
\end{figure*}

\section{Representative construction examples of OmniFake}
\label{app:show}
Figure~\ref{fig:dataset_show} presents representative construction examples from OmniFake. The five columns correspond to \textbf{Real}, \textbf{Human-crafted}, \textbf{Vision Manipulation}, \textbf{Text Manipulation}, and \textbf{Mixed Manipulation}, respectively. Rows (a)--(f) illustrate six representative generation pipelines, including human-written misinformation, full-image synthesis, face swapping, keyword/phrase rewriting, caption rewriting, and object/background replacement. Original samples, intermediate editing procedures, and the resulting manipulated outputs are shown to illustrate how different manipulation types are constructed in OmniFake.

\end{document}


\title{Towards Unified Multimodal Misinformation Detection in Social Media:  A Benchmark Dataset and Baseline}

\maketitle

\section{Introduction}
sup
